\documentclass{bmvc2k}
\usepackage{multirow}
\usepackage{hyperref}

\title{Describe Your Facial Expressions by Linking Image Encoders and Large Language Models}

\addauthor{Yujian Yuan}{yuanyujian18@mails.ucas.ac.cn}{1,2}
\addauthor{Jiabei Zeng}{jiabei.zeng@ict.ac.cn}{1,2}
\addauthor{Shiguang Shan}{sgshan@ict.ac.cn}{1,2}

\addinstitution{
Institute of Computing Technology, Chinese Academy of Sciences\\
Beijing, China
}
\addinstitution{
University of Chinese Academy of Sciences\\
Beijing, China
}

\runninghead{Yuan, Zeng, Shan}{Describe Your Facial Expressions}

\begin{document}

\maketitle

\begin{abstract}

This paper presents a novel task of describing human facial expressions of a facial image in natural language, which captures the nuances of facial actions and emotional states beyond traditional emotion categories or facial action units (AUs). To achieve the facial expression captioning model, we propose a three-stage training framework that trains a vision-to-language model using synthesized image-text pairs and the BLIP-2 pre-training techniques. To overcome the challenge of missing training image-text pairs for facial expression captioning, we propose a strategy that involves synthesizing and combining captions using GPT-3.5 and existing annotations on either emotion categories or AUs. Experiments demonstrate the effectiveness of our method in generating captions that describe details of facial actions and emotions, as well as the inferential relationship between them, even when those emotions are not present in the training data.
It is also demonstrated that the vision-to-language task enhances the performance of the intermediate visual features on both AU detection and emotion classification tasks. The code and trained models are available at: \url{https://github.com/Yujianyuan/Exp-BLIP}.

\end{abstract}
\section{Introduction}
\label{sec:intro}

Facial expressions are a crucial form of nonverbal communication that convey a wide range of emotional states through subtle changes in facial muscle movements. The accurate perception and interpretation of facial expressions are critical for effective human-machine interactions. Existing works have focused on describing facial expressions by classifying them into pre-defined emotion categories (e.g., happiness, sadness, neutral, anger, disgust, surprise, fear) or detecting the appearance of facial action units (AUs) defined by the Facial Action Coding System (FACS)\cite{ekman1978manual}.
However, these methods have limitations. Pre-defined emotion categories cannot capture all the nuances of facial expressions, while AUs do not convey the indicated affective status.
 To overcome these limitations, we propose a novel facial expression captioning task that aims to describe facial actions and the corresponding emotional states of a facial image in natural language. Our approach provides a comprehensive and detailed understanding of facial expressions, which has the potential to improve accuracy and effectiveness in understanding and interacting with humans.

It is non-trivial to accomplish facial expression captioning due to the lack of training image-text pairs that specifically describe nuances of facial expressions. 
Although existing image captioning models such as CoCa~\cite{yu2022coca}, BLIP~\cite{li2022blip}, and Flamingo~\cite{alayrac2022flamingo} demonstrate remarkable performance in describing the contents of an image, they fall short in describing the nuances of facial expressions. This limitation arises from the fact that the training image-text pairs typically do not provide the level of detail required to train a facial expression captioner. 
However, collecting such image-text pairs presents a significant challenge. Annotators must possess exceptional writing skills and the ability to recognize and describe the full range of facial expressions. Additionally, even with expert annotators, the process of annotation is time-consuming.

To overcome the scarcity of training image-text pairs for facial expression captioning, we propose a three-stage framework that leverages synthetic image-text pairs synthesized using GPT-3.5 and existing annotations on emotion categories and AUs. 
As data labeled with both emotion categories and AUs are rarely available, our proposed framework combines captions generated by the emotion-specific captioner (Emot-BLIP) and the AU-specific captioner (AU-BLIP) to train the final captioner (Exp-BLIP) using the combined captions and their corresponding images.
Our contributions are summarised as follows:
\begin{itemize}
\setlength{\itemsep}{0pt}
\setlength{\parsep}{0pt}
\setlength{\parskip}{0pt}
\vspace{-0.8 em}
\item[(1)]
We propose facial expression captioning, a new task to describe nuances of facial expressions and infer the corresponding emotions in words. Compared with conventional emotion classification and AU detection tasks, facial expression captioning interprets facial expressions in a more detailed and comprehensive way.

\item[(2)]
To train the captioner for facial expression captioning, we propose a three-stage framework that utilizes synthetic image-text pairs. This approach enables the training even when the training data that describe both facial actions and emotions are not available.

\item[(3)]
Extensive experiments demonstrate the powerful ability of the trained facial expression captioner and the intermediate visual representation. 
Our method produces captions that describe details of the facial actions and emotions, even if the emotions are not in the training data. 
Experiments also demonstrate that vision-to-language task improves the representation ability of the intermediate visual features on AU detection and emotion classification tasks.
\vspace{-1 em}
\end{itemize}

\section{Related works}
\vspace{-0.4em}
\textbf{Facial expression description method:}
The current approaches to describing facial expressions primarily involve emotion classification\cite{wang2020suppressing,zhang2022learn,zeng2022face2exp,lukov2022teaching} and AU detection\cite{shao2021jaa,jacob2021facial,wang2022action,luo2022learning} tasks. Emotion classification typically consists of categorizing emotions into eight basic categories: neutral, anger, disgust, fear, happiness, sadness, surprise, and contempt~\cite{matsumoto1992more}. To address the limitations of describing facial expressions with single emotions, compound emotions~\cite{du2014compound} was proposed by combining several basic emotions. To overcome the shortcomings of archetypal expression descriptions, Ekman et al.~\cite{ekman1978manual} developed the facial action coding system (FACS) based on facial anatomy, which defines 44 AUs by focusing on different facial parts.
However, it is important to note that both emotions and AUs alone cannot fully capture the complexity of facial status. Recently, Nezami et al.~\cite{nezami2020image} have proposed a novel image captioning model that utilized facial expression features to generate image captions. Their work stated that the improvement in caption quality appeared to come not from the addition of adjectives linked to emotional aspects of the images, but from more variety in the actions described in the captions. Furthermore, Bryson~\cite{lingenfelter2021face} introduced a meticulous approach to generating weak prominent feature labels using semantic segmentation, demonstrating how these labels could enhance attribute-based face descriptions.

\textbf{Application of visual and language pre-trained models:}
Recently, pre-trained vision models and large language models (LLMs) have made remarkable contributions to computer vision and natural language processing (NLP). Numerous studies~\cite{zhou2022learning,lin2022frozen,luddecke2022image,ju2022prompting,zhang2022tip,zhang2023prompt} have demonstrated the efficacy of directly applying the features of these large models to downstream tasks without fine-tuning. This approach has proven impactful in accomplishing multi-modal tasks such as image captioning and vision question answering. Recent multi-modal works~\cite{radford2021learning,jia2021scaling,li2021align,alayrac2022flamingo,bao2022vlmo,yu2022coca,wang2022image} primarily focused on cross-modal alignment. For instance, CLIP\cite{radford2021learning} directly aligned visual and text features during training, while Flamingo\cite{alayrac2022flamingo} and BLIP-2\cite{li2023blip} established a connection between pre-trained vision and language models.

 Prompt engineering is a vital technique in utilizing LLMs. A text or template named prompt can be used to strongly guide the generation to output answers for desired tasks, thus beginning an era of “pre-train and prompt”~\cite{liu2021makes}. Proper zero-shot or few-shot prompts~\cite{brown2020language,min2022noisy,ye2022ontology} have the potential to effectively harness the capabilities of LLMs. For the recent studies, Sewon et al.~\cite{min2022noisy} introduced a noisy channel approach for language model prompting in few-shot text classification. To deal with logical reasoning tasks, Takeshi Kojima et al.~\cite{kojima2022large} proved that step prompting was a more powerful prompt than zero-shot prompts. 

\begin{figure}[t]
    \centering
    
    \includegraphics[width=1.0\textwidth]{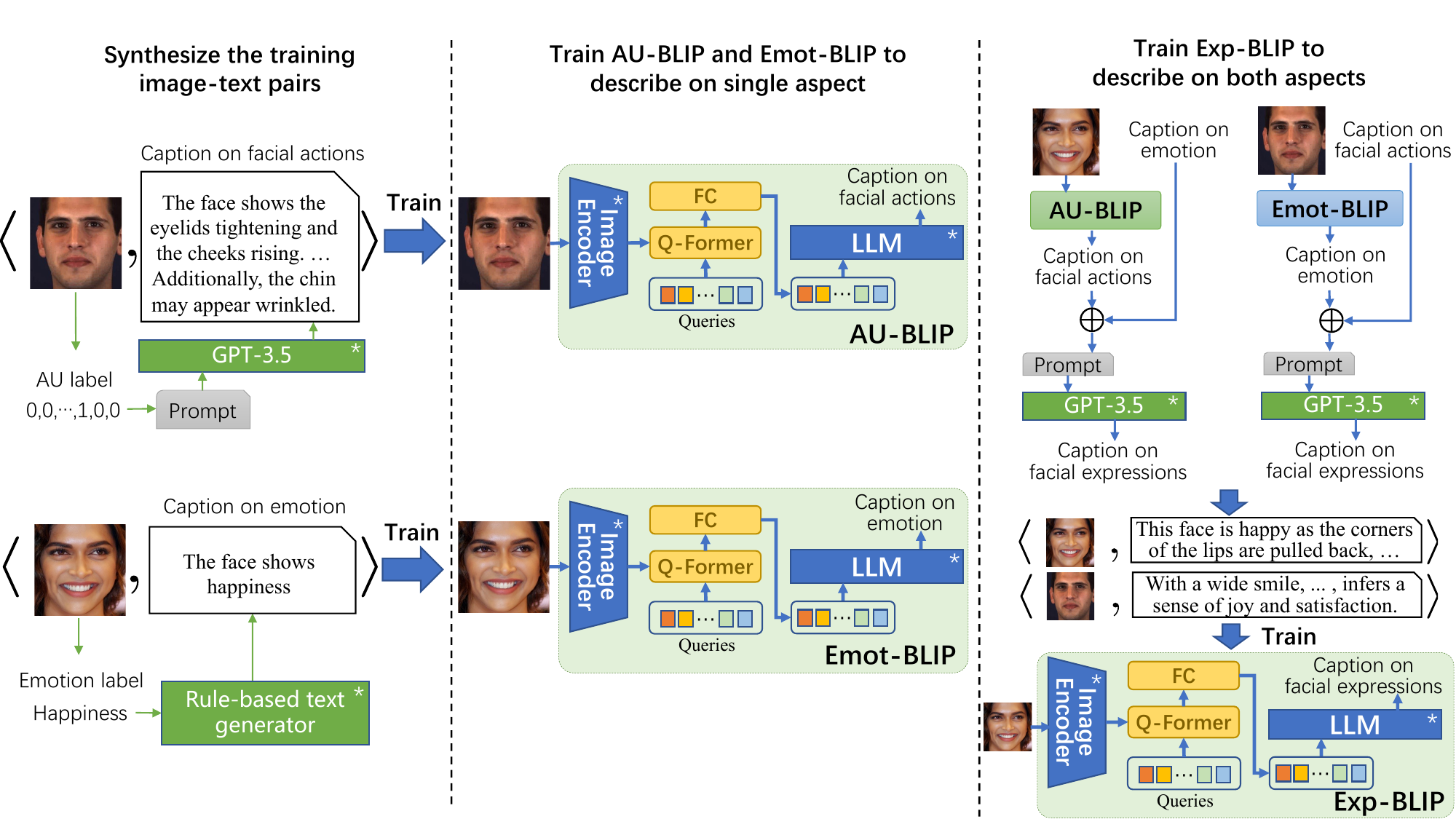}
    \vspace{-2 em}
    \caption{Three-stage training framework for facial expression captioning. 
    \textbf{(Left)} Synthesize the training image-text pairs. \textbf{(Middle)} Train AU-BLIP and Emot-BLIP to respectively describe a single aspect of facial expression. \textbf{(Right)} Train Exp-BLIP utilizing synthesized data with fused captions to describe both facial actions and emotions.
    }
    \label{fig:overview_structure}
    \vspace{-0.5 em}
\end{figure}

\vspace{-0.9 em}
\section{Facial Expression Captioning}
\vspace{-0.6 em}
This paper aims to develop an image captioner that describes the nuances of facial actions and the corresponding emotional states of a facial image in natural language. 
To achieve this goal, we propose a three-stage framework that leverages synthetic image-text pairs. 
Figure~\ref{fig:overview_structure} illustrates the proposed framework. 
In the first step, we synthesize the training image-text pairs using the prevalent large language model (e.g., GPT-3.5) or a rule-based text generator.
In the second step, we train two separate captioners, AU-BLIP and Emot-BLIP, using synthetic pairs that describe facial expressions' specific aspects, including AUs and emotions. 
We do not directly train an image captioner that simultaneously describes both facial actions and emotions, as it is a challenge to derive training texts that correctly describe both two aspects due to few datasets being annotated with both emotions and AUs. 
In the third step, we combine the outputs of AU-BLIP and Emot-BLIP to synthesize descriptions for each image. With the combined captions, we train the final captioner (Exp-BLIP) to describe both aspects of facial expressions. 
Below, we present details of the three steps.

\subsection{Synthesis of the training image-text pairs}\label{step1}

\begin{figure}[t]
    \centering
    \includegraphics[width=1.0\textwidth]{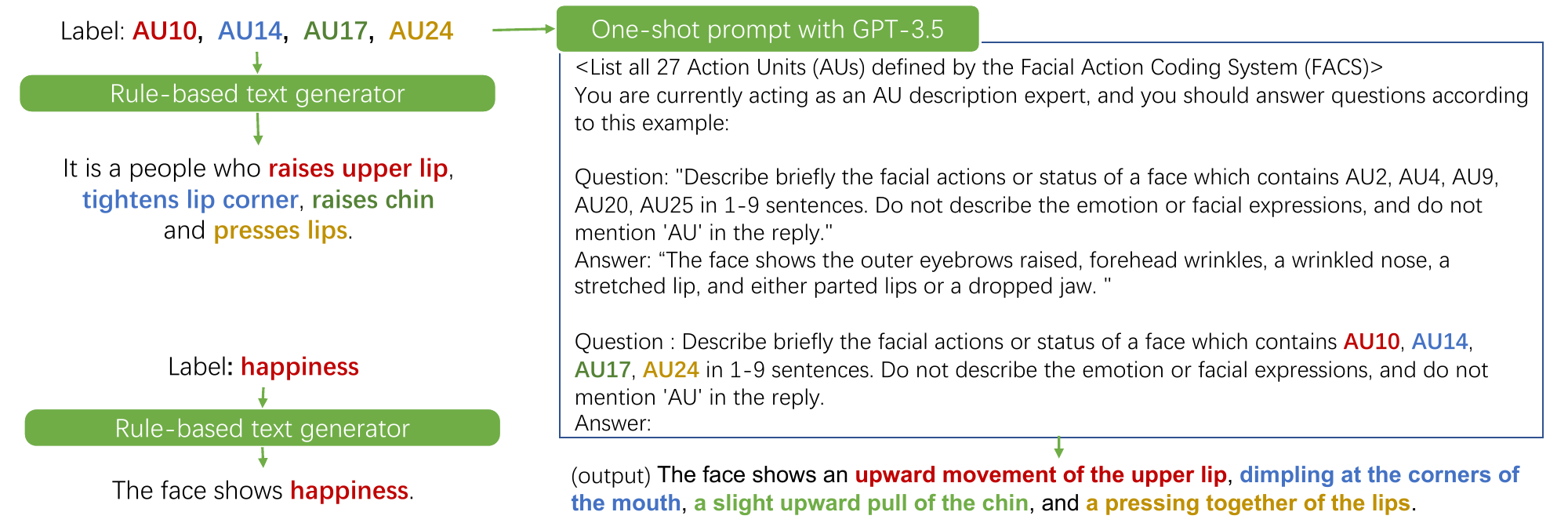}
    \vspace{-2 em}
    \caption{Inputs and outputs examples of the rule-based text generators for AUs and emotions(Left) and those of the GPT-3.5 based generator with one-shot prompt(Right) for AUs.}
    \label{fig:prompt_method}
    \vspace{-1 em}
    
\end{figure}

To train the image captioners, AU-BLIP and Emot-BLIP that are respectively specialized for AUs and Emotions, we constitute the training image-text pairs by synthesizing descriptive texts based on annotations of AUs and emotion categories for the corresponding facial images.
The texts are synthesized in two ways: the rule-based text generator, and the GPT-based generator with one-shot prompt.
Examples of the two types of generators are shown in Figure~\ref{fig:prompt_method}. We use both generators to ensure diverse descriptions of facial actions. For synthesizing descriptions of emotions, we only use the rule-based text generator.

\textbf{Rule-based labeling for AU captions:} 
To synthesize concise and complete descriptions of facial actions, we leverage the annotations of AUs because most facial muscle movements are depicted as AUs by FACS\cite{ekman1978manual}. 
As shown in Fig.~\ref{fig:prompt_method}, given a facial image and its label that indicates the appearance of each AU, we combine the brief descriptions of the appearing AU and make a sentence in the form of "It is a people who <action in appearing AUs>". 
The brief descriptions of each AU are listed in the supplemental materials. %
If no AUs are labeled as appearing in a face, we do not synthesize rule-based AU descriptions for the image.

\textbf{GPT-based labeling for AU captions:}
To ensure diverse training texts, we also utilize GPT-3.5 to synthesize descriptions for facial actions. 
However, the outputs of GPT-3.5 are uncontrolled and it might include the re-introduction of AUs, the induction of different emotions, and other irrelevant contents.
To achieve a clear and concise description, we design a one-shot prompting strategy that requires GPT-3.5 to act as an AU expert and provides it with an example. 
Fig.~\ref{fig:prompt_method} shows an example. We first claim the list of AUs' definitions and then state the requirement through a Question-Answer example.
We explicitly add two restrictions to regulate GPT-3.5's outputs, prohibiting the induction of emotion and the repetition of the word "AU" in the reply.
Then, we throw out the question that asks GPT-3.5 to describe the face with the annotated AUs. If no AUs are labeled, we use a predetermined description to avoid inaccurate outputs from GPT-3.5, which is listed in the supplemental material.

\textbf{Rule-based labeling for emotion captions:} 
To keep the correct description of emotions, %
we synthesize the descriptions of emotions by simply generating a sentence in the form of "The face shows <emotion category>" according to the manual annotations in the existing datasets. 
If the images are labeled with more than one emotion category, e.g., compound emotions\cite{li2017reliable} or multi-label emotion category\cite{lu2019zero}, we merge the involved emotion labels.

\vspace{-0.5 em}
\subsection{Training of AU-specific and emotion-specific image captioners}\label{step2}
\vspace{-0.4em}

We use the synthetic pairs in Sec.~\ref{step1} to train two image captioners:
One describes facial actions (AU-BLIP),
and the other describes emotions (Emot-BLIP). 
Both two captions are trained following the framework in BLIP-2\cite{li2023blip}.
In addition, other vision-to-language models can also be adopted in our three-stage facial expression captioning framework.

\begin{figure}[t]
    \centering
    \includegraphics[width=1.0\textwidth]{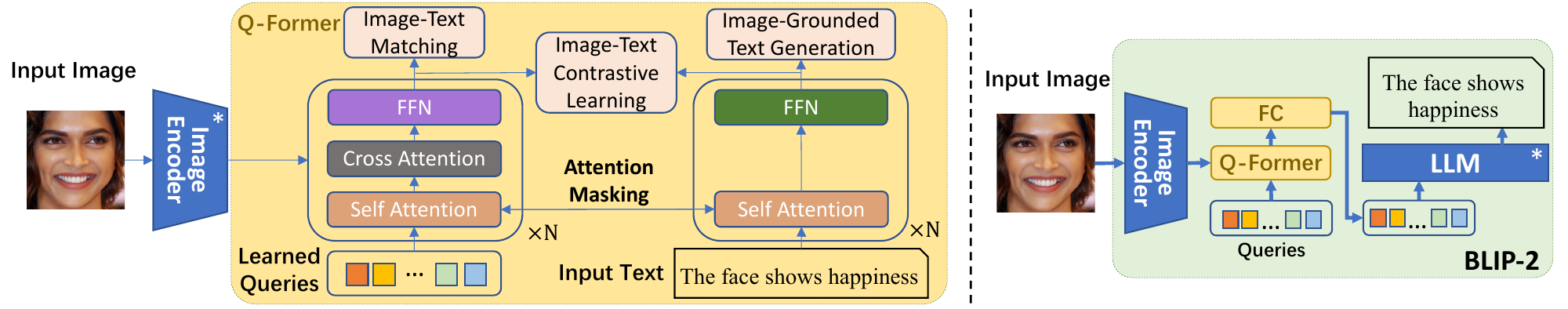}
    \caption{Two-step training procedure of BLIP-2\cite{li2023blip}. The 1$^\text{st}$ step (\textbf{Left}): BLIP-2 enforces the queries to extract the visual representation most relevant to the text. The 2$^\text{nd}$ step(\textbf{Right}): BLIP-2 bootstraps from frozen LLMs to conduct vision-to-language generative training.}
    \label{fig:BLIP2_structure}
    \vspace{-0.5 em}
\end{figure}

\textbf{Revisiting BLIP-2:} Fig.~\ref{fig:BLIP2_structure} illustrates the architecture of BLIP-2 and its two-step training procedure. 
As can be seen in Fig.~\ref{fig:BLIP2_structure} (Right), BLIP-2 takes an image as input, and then extracts the visual features from a pre-trained image encoder.
The visual features are then passed through a Q-Former and a pre-trained large language model(LLM). %
Finally, BLIP-2 outputs a description of the input image.
Q-Former is a trainable module that connects the image encoder and LLM. The detailed structure of Q-Former is illustrated in Fig.~\ref{fig:BLIP2_structure}(Left). 
It is trained in two steps.
In the first step, as shown in Fig.~\ref{fig:BLIP2_structure}(Left), to guide the learned queries to aggregate the vision-language representation from the image encoder and training texts, BLIP-2 freezes the image encoder and optimizes Q-Former by minimizing the Image-Text Contrastive Learning (ITC) loss, Image-grounded Text Generation (ITG) loss and Image-Text Matching (ITM) loss.
The self-attention layers for the queries and the texts share the same parameters.
As BLIP-2\cite{li2023blip} does, attention masking is applied to prevent information leakage. 
In the second step, to harvest the LLM’s generative language capability and adjust the image encoder to extract features that are more related to the image captioning task, BLIP-2 connects Q-Former to a frozen LLM, as shown in Fig.~\ref{fig:BLIP2_structure} (Left), and jointly update the parameters of Q-Former and the image encoder by 
minimizing the discrepancy between the LLM-generated caption and the ground truth.

\textbf{Pre-trained image encoder and LLM:} For the image encoder, we utilize ViT-G/14 from EVA-CLIP~\cite{fang2022eva} and a ViT-Base trained on AffectNet~\cite{mollahosseini2017affectnet} dataset under the training framework of MAE~\cite{he2022masked}. Consistent with BLIP-2, we choose the unsupervised-trained OPT model family~\cite{zhang2022opt} as our language model. Model details are listed in supplementary material.

\begin{figure}[t]
    \centering
    \includegraphics[width=1.0\textwidth]{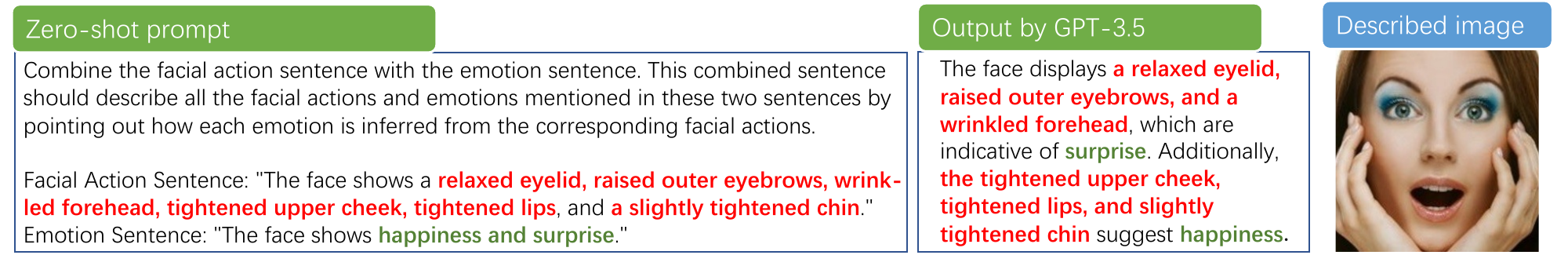}
    \vspace{-2.2 em}
    \caption{Inputs and outputs examples of the GPT-3.5 based generator with zero-shot prompt for fusing AU and emotion captions. The image described by the fused caption is on the right.}
    
    \label{fig:prompt_zero_shot}
    \vspace{-0.8 em}
    
\end{figure}

\vspace{-1.6 em}
\subsection{Exp-BLIP with fused captions}\label{fuse}
\vspace{-0.6 em}
We train the final facial expression captioner (Exp-BLIP) with the synthesized descriptions by GPT-3.5 and AU-BLIP/Emot-BLIP.
Fig.~\ref{fig:overview_structure} (Right) shows how we train Exp-BLIP.
To assign each training image a description of both AU and emotion, we use AU-BLIP/Emot-BLIP to generate a pseudo-caption if it lacks the original annotation on AU/Emotion.
Then, we utilize GPT-3.5 to fuse the two captions with zero-shot prompting, leveraging the powerful inferring ability of GPT-3.5. 
Fig.\ref{fig:prompt_zero_shot} shows an example of zero-shot prompting in fusing captions.
With the fused captions, we train Exp-BLIP following the architecture and training procedure of BLIP-2\cite{li2023blip} in Sec.~\ref{step1}.

\begin{table}[t]
    \caption{Statistics of training and test data.(\#sub:number of subjects; *:sampled set)}
    \label{tab:datasets}
    \centering
    \footnotesize%
    \setlength{\tabcolsep}{2pt}%
    \renewcommand{\arraystretch}{1.2}%
    \begin{tabular}{l|ccccc|ccc}
        \hline
        \multirow{2}{*}{} & \multicolumn{5}{c|}{Originally annotated with AUs} & \multicolumn{3}{c}{Originally annotated with emotions}      \\
          &  BP4D & DISFA & GFT & RAF-AU & EmotioNet &  AffectNet & RAF-DB & FaceME   \\
        \hline
        train($\#$image/$\#$sub) & $16627^*$/28 & $14814^*$/24 & $17719^*$/78 & 3733/- & 19046/- & 287618/- & 3162/- & 10052/- \\
        test($\#$image/$\#$sub) & 45805/13 & 14535/3 & $4034^*$/18 & 868/- & 2117/- & 4000/- & 792/- & - \\
        \hline
    \end{tabular}
    \vspace{-1 em}
\end{table}

\vspace{-1.6 em}
\section{Experiment}
\vspace{-1 em}
We evaluate the performance of AU-BLIP, Emot-BLIP, and Exp-BLIP from two perspectives: the quality of generated captions, and the ability of visual representation.

\vspace{-1.3em}
\subsection{Experiment settings}
\vspace{-0.7em}
\textbf{Training data:} 
We used nearly 372k training image-text pairs in total. 
72k of the training data are derived from AU datasets BP4D~\cite{zhang2013high}, DISFA~\cite{mavadati2013disfa}, GFT~\cite{girard2017sayette}, RAF-AU~\cite{yan2020raf} and EmotioNet~\cite{fabian2016emotionet} using the rule-based text generator and GPT-based generator in Sec.~\ref{step1}. Texts from the two generators were both used to train AU-BLIP. Only the GPT-3.5 generated ones were used to synthesize training pairs for Exp-BLIP. 
Considering the high cost of using GPT-3.5 API\footnote{\url{https://platform.openai.com/docs/guides/gpt/chat-completions-api}} and the significant similarity among the consecutive frames in video-based AU datasets, we first select one sample from every ten frames. Then, we selected the frames which had different AU labels compared with the previous frame to fully utilize the datasets. This selection was conducted in training sets of video-based AU datasets (BP4D, DISFA, and GFT) and test set of GFT.
300k of the training data are derived from AffectNet, RAF-DB~\cite{li2017reliable} and FaceME~\cite{lu2019zero}.
All of the datasets except FaceME are split into training and test parts without overlapped subjects. 
Tab.~\ref{tab:datasets} shows the statistics of training and test data. 

\textbf{Evaluation protocol for image captioning:} We evaluated our image captioning models in two ways: (1) computing metrics for AU captioning, and (2) conducting manual evaluation to assess the completeness and correctness of facial expression captioning.
The former was conducted on both in-the-lab data (test set of GFT) and in-the-wild data (test set of RAF-AU).
we chose Meteor~\cite{banerjee2005meteor}, Rouge~\cite{lin2004rouge} and Ada-similarity (Ada-sim) between the generated captions and the ground truths as our metrics. 
Ada-similarity calculated the cosine similarity of two features extracted from text-embedding-ada-002\footnote{\url{https://platform.openai.com/docs/guides/embeddings/what-are-embeddings}}. 
Since each sample has multiple ground truths (synthesized by GPT-3.5 and the rule-based generator), we computed the metrics between the predicted caption and each ground truth, and then we chose the highest score.
The latter was conducted on 505 samples randomly chosen from all the test data in Tab.\ref{tab:datasets}.
Then, we invited 9 human labelers to rate the correctness of each predicted caption from 1 to 10 and to choose the most/least complete one from the three predictions.

\textbf{Evaluation protocol for visual presentation:} 
We evaluated the visual representations from the image encoder and Q-Former by investigating their performance in AU detection and emotion classification tasks using a linear probe or a fine-tuning strategy. 
We added a fully connected layer after the representation as the classifier. The AU classifier was trained on full BP4D and RAF-AU training set without sampling, while the emotion classifier was trained on AffectNet. %
In the linear probe, the image encoder and Q-Former were frozen.
In the fine-tuning strategy, the parameters of the image encoder and Q-Former were updated. 
F1 score and classification accuracy were used to measure the performance of AU detection and emotion classification, respectively.
To compare with state-of-the-art (SOTA) method in RAF-AU, we selected the same AUs of AU-CNN\cite{yan2020raf} for training.

\textbf{Implementation:} During the training of AU-BLIP, Emot-BLIP, and Exp-BLIP, we used a batch size of 28 for the first step and 32 for the second step. We used the AdamW~\cite{loshchilov2017decoupled} optimizer and a weight decay of 0.05. A cosine learning rate decay with a peak learning rate of 1e-4 and a linear warm-up of 2k steps was adopted. The minimum learning rate in the second step was 5e-5. The input images were of size 224×224, augmented with random resized cropping and horizontal flipping. All the experiments run on 4 Nvidia A100 (40G). %

\begin{table}[t]
    \caption{Metrics of AU captioning with different models (ViT-B/G denotes the used image encoder, 2.7B/6.7B denotes the used language model OPT-2.7B/6.7B)}
    \label{tab:AU caption}
    \centering
    \footnotesize%
    \setlength{\tabcolsep}{3pt}%
    \renewcommand{\arraystretch}{1.2}%
    \begin{tabular}{l|ccc|ccc}
        \hline
        \multirow{2}{*}{Models} & \multicolumn{3}{c|}{RAF-AU} & \multicolumn{3}{c}{GFT}      \\
          &  Meteor $\uparrow$ & Rouge $\uparrow$ & Ada-sim($\%$) $\uparrow$ &  Meteor $\uparrow$ & Rouge $\uparrow$ & Ada-sim($\%$) $\uparrow$    \\
        \hline
        BLIP-2(ViT-G,6.7B) & 0.031 & 0.112 & 76.32 & 0.033 & 0.118 & 77.58 \\
        BLIP-2(ViT-G,6.7B)$^\text{COCO}$ & 0.034 & 0.129 & 77.31 & 0.034 & 0.131 & 78.47\\
        \hline
        AU-BLIP(ViT-B,2.7B)      & 0.236 & 0.457 & 90.13 & 0.278 & 0.509 & 93.11 \\
        AU-BLIP(ViT-G,2.7B)    & 0.254 & 0.478 & 90.23 & 0.279 & 0.487 & 93.45\\
        AU-BLIP(ViT-G,6.7B)    & \textbf{0.263} & \textbf{0.505} & \textbf{90.55}  & \textbf{0.295} & \textbf{0.538} & \textbf{93.97}\\
        \hline
        Mix-BLIP(ViT-B,2.7B)     & 0.158 & 0.324 & 86.00 & 0.273 & 0.459 & 92.55 \\
        Mix-BLIP(ViT-G,6.7B)   & 0.204 & 0.364 & 87.87  & 0.279 & 0.489 & 93.00\\
        Cat-BLIP(ViT-B,2.7B)  & 0.211 & 0.374 & 87.38 & 0.240 & 0.355 & 92.02 \\
        Cat-BLIP(ViT-G,6.7B)  & 0.235 & 0.426 & 89.76 & 0.248 & 0.385 & 92.88 \\
        Exp-BLIP(ViT-B,2.7B)     & 0.164 & 0.225 & 83.86 & 0.215 & 0.279 & 90.72 \\
        Exp-BLIP(ViT-G,6.7B)   & 0.184 & 0.229 & 84.65  & 0.217 & 0.318 & 90.75\\
        \hline
    \end{tabular}
    \vspace{-1.5 em}
\end{table}

\vspace{-0.8em}
\subsection{Experiment results and analysis}
\begin{figure}[!htbp]
    \centering
    \includegraphics[width=1.0\textwidth]{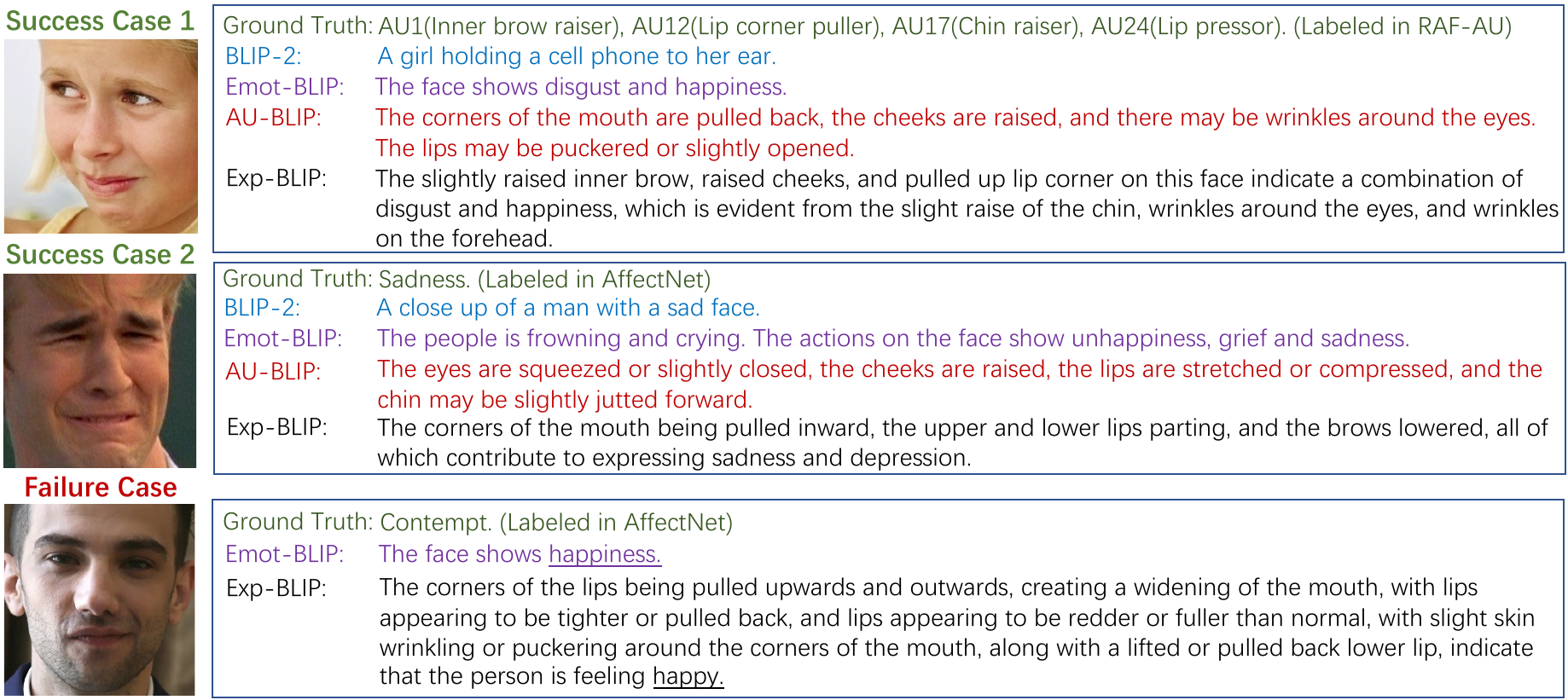}
    \vspace{-2 em}
    \caption{Examples of the generated captions by different models with ViT-G/OPT-6.7B}
    \label{fig:examples}
    \vspace{-0.2 em}
\end{figure}

\vspace{-0.4 em}
\textbf{Evaluation on image captioning:}
Table~\ref{tab:AU caption} reports the metrics on AU captioning using the original BLIP-2\cite{li2023blip}, AU-BLIP, Exp-BLIP, Mix-BLIP, Cat-BLIP with different versions of image encoder and LLMs. 
Mix-BLIP and Cat-BLIP adopt different approaches to fuse AU caption and emotion caption(Sec.~\ref{ablation}).
BLIP-2 is the model provided by \cite{li2023blip} that was trained on 129M images.
BLIP-2$^\text{COCO}$ is the fine-tuned BLIP-2 on COCO~\cite{lin2014microsoft} for general image captioning.
As shown in Table~\ref{tab:AU caption}, our models trained on AU-specific data achieve consistently larger values than the two models trained for general image captioning, because the general image captioner describes limited information on facial expressions, as shown in Fig.~\ref{fig:examples}.
It indicates the necessity to train a task-specific vision-to-language model for describing facial expressions even though the general captioning model has shown powerful ability.

It is also observed in Table~\ref{tab:AU caption} that AU-BLIP has larger values in metric than the fused-caption models. 
It is because the ground truths are synthetic AU captions, while Mix-BLIP tends to randomly generate AU or emotion captions, and Cat/Exp-BLIP describe extra information about emotions, which degrades the metrics to AU captioning.

Figure~\ref{fig:examples} shows examples of the generated captions by different models. 
All the models are with image encoder ViT-G and language model OPT-6.7B.
As can be seen, BLIP-2 describes very few facial expressions but describes other attributes of faces.
Emot-BLIP describes emotions with limited facial actions.
AU-BLIP only describes facial actions.
Exp-BLIP not only describes both facial actions and emotions, but also describes the inferential relation between them. 
Our models produce inaccurate results in some cases.
For example,
in the failure case of figure~\ref{fig:examples}, our Emot-BLIP and Exp-BLIP infer different emotions from the ground truth.
Although labeled as contempt, this picture also slightly shows happiness as we predict.
More failure cases for different reasons are listed in the \textit{suppl. material}.

\vspace{-0.5 em}

\textbf{Evaluation on visual representation:} 
Tables \ref{tab:BP4D eval} and \ref{tab:RAFAU eval} report the performance of visual representations, including F1 scores on BP4D and RAF-AU datasets, and classification accuracy on AffectNet dataset. 
In addition, the comparisons with SOTA facial representation (FaRL\cite{zheng2022general}), AU detectors (Me-GraphAU\cite{luo2022learning}, AU-CNN\cite{yan2020raf}), emotion classifier (EfficientNet-B2\cite{savchenko2022classifying}) are included.
ViT-G stands for the pre-trained image encoder mentioned in Sec.~\ref{step2}. The values are reported under a linear probe strategy.
ViT-G$^\text{fine-tune}$ is ViT-G fine-tuned on the training data for AU detection or emotion classification, using a fine-tuning strategy.
Other models with the names formatted as <ViT-G/QFormer>(AU/Emot/ Mix/Cat/Exp) denote the visual representations (e.g., the image encoder or Q-Former) from the varied models (i.e., AU-BLIP, Emot-BLIP, Mix-BLIP, Cat-BLIP, Exp-BLIP) with LLM OPT-6.7B.
For these representations and FaRL, linear probe strategies were applied. Other AU detectors and emotion classifiers used the values reported in their papers.
It is noting that Me-GraphAU was evaluated under a 3-fold protocol on BP4D, while our models were evaluated on one of the three folds, the one that was not used in training AU/Exp-BLIPs.
All the models were trained for 100 epochs on BP4D and 400 epochs on RAF-AU.

\begin{table}[t]
    \caption{Performance of visual representation on BP4D and AffectNet. ($^*$: original values)}
    \label{tab:BP4D eval}
    \centering
    \scriptsize%
    \setlength{\tabcolsep}{2pt}%
    \renewcommand{\arraystretch}{1.2}%
    \begin{tabular}{lccccccccccccc|c}
        \hline
        \multirow{2}{*}{Models} & \multicolumn{13}{c|}{BP4D (F1 score$\times 100$)} & AffectNet (\textit{Acc.}\%)\\
        & AU1& AU2& AU4& AU6& AU7& AU10& AU12& AU14& AU15& AU17& AU23& AU24 & Avg & Emotion\\
        \hline
        FaRL\cite{zheng2022general}         & 48.4 & 45.9 & 47.5 & 78.9 & 69.8 & 82.8 & 86.0 & 58.3 & 42.0 & 55.9 & 36.1 & 38.0 & 57.5 & 38.8 \\
        Me-GraphAU$^*$\cite{luo2022learning}   & 52.7 & 44.3 & 60.9 & 79.9 & \textbf{80.1} & 85.3 & 89.2 & \textbf{69.4} & \textbf{55.4} & 64.4 & \underline{49.8} & 55.1 & 65.5 & - \\
        EfficientNet-B2$^*$\cite{savchenko2022classifying} & - & - & - & - & - & - & - & - & - & - & - & - & - & \textbf{63.0} \\
        \hline
        ViT-G              & 53.2 & 34.2 & 58.7 & 81.1 & 71.1 & 85.7 & 90.0 & 56.1 & 31.2 & 58.3 & 43.4 & 51.1 & 59.5 & 44.8\\
        ViT-G$^\text{fine-tune}$ & 32.8 & 18.4 & 38.1 & 77.2 & 68.6 & 84.0 & 85.1 & 60.1 & 32.3 & 61.6 & 35.7 & 32.9 & 52.2  & 50.3\\
        \hline
        ViT-G(Emot)  & 44.1 & 28.7 & 54.0 & 80.5 & 76.2 & \underline{88.1} & 90.5 & \underline{64.9} & 44.9 & 55.2 & 43.6 & 50.5 & 60.1  & 50.9\\
        QFormer(Emot)  & 43.7 & 21.0 & 54.8 & 80.7 & \underline{78.8} & 87.6 & \textbf{91.4} & 58.7 & 43.4 & 52.7 & 34.8 & 52.1 & 58.3  & \underline{52.7} \\
        ViT-G(AU)    & 57.7 & \underline{54.8} & \textbf{67.0} & 81.1 & 75.6 & 86.6 & 90.1 & 60.2 & \underline{51.8} & \underline{64.6} & 49.4 & 56.8 & \underline{66.3}  & 50.1\\
        QFormer(AU)    & \textbf{59.9} & \textbf{55.9} & \underline{65.7} & 81.2 & 76.7 & 87.7 & 90.8 & 60.4 & 50.8 & \textbf{64.8} & 49.4 & 55.7 & \textbf{66.6} & 52.2 \\
        ViT-G(Mix)   & 59.0 & 47.6 & 62.3 & 80.8 & 76.3 & 86.4 & 87.9 & 64.0 & 45.6 & 57.2 & 44.3 & 53.3 & 63.7   & 48.2 \\
        QFormer(Mix)   & 58.7 & 46.3 & 61.7 & \underline{81.5} & 77.0 & 85.1 & 90.2 & 64.2 & 43.7 & 59.0 & 45.2 & 55.7 & 64.0  & 49.3 \\
        ViT-G(Cat)  & 58.9 & 46.9 & 60.7 & 80.8 & 76.9 & 87.2 & 88.0 & 61.5 & 45.3 & 60.3 & 43.0 & 55.6 & 63.8    & 48.9 \\
        QFormer(Cat)  & 57.8 & 48.4 & 57.3 & \textbf{82.1} & 77.2 & 85.3 & 90.6 & 64.4 & 44.0 & 58.2 & 46.1 & 55.3 & 63.9  & 49.5 \\
        ViT-G(Exp)  & 59.2 & 51.7 & 62.5 & 80.9 & 78.1 & \underline{88.1} & 90.2 & 63.4 & 48.5 & 62.8 & 47.4 & \textbf{61.7} & 66.2   & 51.0 \\
        QFormer(Exp)  & \textbf{59.9} & 52.9 & 61.3 & 81.3 & 78.7 & \textbf{88.2} & \underline{90.7} & 64.5 & 47.6 & 61.7 & \textbf{50.5} & \underline{57.0} & 66.2  & 51.6 \\
        \hline
    \end{tabular}
    \vspace{-1.3 em}
\end{table}

\begin{table}[!htbp]
    \caption{Performance of visual representation on RAF-AU. ($^*$: original values)}
    \label{tab:RAFAU eval}
    \centering
    \scriptsize%
    \setlength{\tabcolsep}{3.5pt}%
    \renewcommand{\arraystretch}{1.2}%
    \begin{tabular}{lcccccccccccccc}
        \hline
        Models & AU1& AU2& AU4& AU5& AU6& AU9& AU10& AU12 & AU16& AU17 & AU25& AU26& AU27& Avg    \\
        \hline
        FaRL\cite{zheng2022general}         & 57.8 & 56.3 & 75.2 & 56.2 & 32.8 & 50.2 & 61.5 & 68.9 & 49.7 & 38.8 & 84.4 & 55.0 & 60.6 & 57.5 \\
        AU-CNN$^*$\cite{yan2020raf}       & 60.5 & 65.6 & 73.4 & 69.7 & \textbf{58.2} & 67.4 & 68.4 & 69.6 & \underline{59.4} & 25.6 & 92.1 & 64.7 & \textbf{82.7} & 65.9 \\
        \hline
        ViT-G             & 60.0 & 57.6 & 77.2 & 62.3 & 35.8 & 59.7 & 70.6 & 73.6 & 53.6 & 45.5 & 88.7 & 56.3 & 70.1 & 62.4    \\
        ViT-G$^\text{fine-tune}$ & 27.6 & 34.5 & 59.1 & 52.5 & 23.4 & 38.5 & 48.9 & 51.0 & 41.9 & 25.0 & 86.1 & 46.5 & 62.6 & 46.0  \\
        \hline
        ViT-G(Emot)   & 61.0 & 55.3 & 78.4 & 62.2 & 37.5 & 66.8 & 70.8 & 71.5 & 53.8 & 48.6 & 88.8 & 55.9 & 69.5 & 63.1  \\
        QFormer(Emot) & 57.8 & 53.6 & 78.8 & 62.8 & 36.1 & 62.9 & 67.2 & 68.4 & 52.7 & 42.4 & 86.6 & 56.1 & 62.0 & 60.6  \\
        ViT-G(AU)    & 74.9 & 68.2 & 82.3 & \textbf{73.9} & \underline{50.2} & \textbf{79.0} & 71.1 & 74.0 & 58.6 & \underline{59.4} & 91.5 & 68.0 & 78.0 & \underline{71.5}  \\
        QFormer(AU)    & \underline{75.2} & \textbf{70.9} & 82.2 & 71.7 & 47.2 & \underline{76.1} & \textbf{73.2} & 74.3 & 59.3 & \textbf{61.4} & 95.6 & \textbf{72.3} & \underline{79.1} & \textbf{72.2}\\
        ViT-G(Mix)   & 69.1 & 60.5 & \textbf{85.2} & 68.7 & 43.8 & 69.9 & 69.0 & 72.9 & 56.9 & 47.7 & 94.5 & 61.5 & 66.7 & 66.6  \\
        QFormer(Mix)   & 74.9 & 64.6 & 83.8 & 69.6 & 47.8 & 72.1 & 68.6 & \underline{74.7} & \textbf{59.6} & 53.5 & 95.4 & 65.5 & 69.8 & 69.2  \\
        ViT-G(Cat)   & 71.8 & 64.4 & 84.6 & 70.4 & 39.2 & 70.0 & 67.9 & 74.2 & 56.0 & 49.4 & \underline{95.8} & 60.9 & 65.8 & 67.0  \\
        QFormer(Cat)  & \textbf{76.8} & 67.1 & \textbf{85.2} & \underline{72.4} & 44.1 & 72.3 & 66.6 & \textbf{75.0} & 57.6 & 58.4 & \textbf{96.2} & 62.3 & 68.7 & 69.4 \\
        ViT-G(Exp)   & 72.6 & 65.4 & 82.5 & 68.9 & 40.8 & 70.6 & \underline{72.9} & 74.0 & 57.6 & 52.4 & 92.3 & \underline{70.1} & 78.8 & 69.1 \\
        QFormer(Exp)  & 74.0 & \underline{68.5} & 82.5 & 70.3 & 42.9 & 74.9 & \underline{72.9} & 73.1 & 57.4 & 52.8 & 93.5 & 68.9 & 71.7 & 69.5\\
        \hline
    \end{tabular}
    \vspace{-1.2 em}
\end{table}

Tables \ref{tab:BP4D eval} and \ref{tab:RAFAU eval} show that most of the ViT-G and Q-Former models with linear probe strategy in our approach achieve higher F1 scores than ViT-G and ViT-G$^\text{fine-tune}$. This suggests that incorporating language tasks improves the visual representations and enhances performance on related downstream tasks.
It is also observed that the features of Q-Former are superior to those of the corresponding image encoder, except for Emot-BLIP for AU detection.
This may be due to Q-Former's ability to encode extra discriminative information about facial actions or emotions through the learned queries.
Q-Former(Emot) has a larger F1 score than ViT-G(Emot). 
This may be because Q-Former(Emot)'s features align more closely with its emotional semantics, resulting in reduced representation ability for capturing AU details. 
It is worth mentioning that ViT-G$^\text{fine-tune}$ performs less powerful than ViT-G on BP4D and RAF-AU, potentially because ViT-G$^\text{fine-tune}$ suffers from over-fitting. When trained on larger dataset (i.e., AffectNet), ViT-G$^\text{fine-tune}$ shows better performance.

It is also observed that AU-BLIP and Emot-BLIP reach the best performance respectively on AU detection and emotion classification.
They outperform all the models with combined AUs and emotions.
One possible reason is that the representation from AU-BLIP and Emot-BLIP contains the least irrelevant information, making it easier to fit a linear classifier.

In the AU detection task,
the performance of QFormer(AU/Exp) is compatible to SOTA facial representation and AU detectors. 
However, in the emotion classification task,
there is a gap between the performance of QFormer(Emot) and SOTA emotion classifier (i.e., EfficientNet-B2), which is worth further exploring.
This may be attributed to the simpleness of synthetic emotion captions and the limited ability of the LLM (OPT-6.7B), resulting QFormer(Emot) benefiting little from the emotion captions.

\subsection{Ablation study} \label{ablation}
\vspace{-0.5em}

\begin{table}[t]
    \caption{Correctness and completeness for different combining strategies}
    \label{tab:mannual score}
    \centering
    \footnotesize%
    \setlength{\tabcolsep}{4pt}%
    \renewcommand{\arraystretch}{1.2}%
    \begin{tabular}{lcccccc}
        \hline
        Models & Correctness $\uparrow$ & The highest completeness(ratio) $\uparrow$ & The lowest completeness(ratio) $\downarrow$\\
        \hline
        Mix-BLIP(ViT-B,2.7B)     & \textbf{7.61} & 3.43\% & 91.17\%\\
        Cat-BLIP(ViT-B,2.7B)  & 7.27 & 28.85\% & 7.91\%\\
        Exp-BLIP(ViT-B,2.7B)     & \textbf{7.61} & \textbf{67.72\%} & \textbf{0.92\%}\\
        \hline
    \end{tabular}
    \vspace{-1.5 em}
\end{table}

\textbf{Strategies for combining AU and emotion captions:} 
Table~\ref{tab:mannual score} displays the assessments of the correctness and completeness of captions generated by three models, which fuse AU and emotion captions using different strategies.
Mix-BLIP trains the facial expression captioner using the mixed data that with only AU captions or emotion captions.
Cat-BLIP differs from Exp-BLIP by directly concatenating the two captions rather than fusing them with GPT-3.5 as Exp-BLIP does (Sec.~\ref{fuse}).
Mix-BLIP achieves the highest correctness while the least completeness among the three methods, which betrays our motivation of generating detailed facial descriptions. 
In Table~\ref{tab:AU caption}, Mix-BLIP and Cat-BLIP have higher metrics for AU captioning than Exp-BLIP, which may be explained by the different formats of training texts. The AU captions in the training data of Mix-BLIP and Cat-BLIP are synthesized by rules, the same as the ground truths used in computing the metrics.
Exp-BLIP is trained with the descriptions generated by GPT-3.5. 
In Table~\ref{tab:BP4D eval} and \ref{tab:RAFAU eval}, the visual representations from Exp-BLIP show superior performance than those from the other two, indicating the advantage of our proposed combination strategy.

\textbf{Scaling language and visual model:} The comparison between methods with (ViT-B, OPT-2.7B), (ViT-G, OPT-2.7B), and (ViT-G, OPT-6.7B) in Table~\ref{tab:AU caption} suggests that in AU captioning, a stronger image encoder or a stronger LLM both lead to better performance, which is consistent with the conclusion in \cite{li2023blip}.

\begin{figure}[!htbp]
    \centering
    \includegraphics[width=1.0\textwidth]{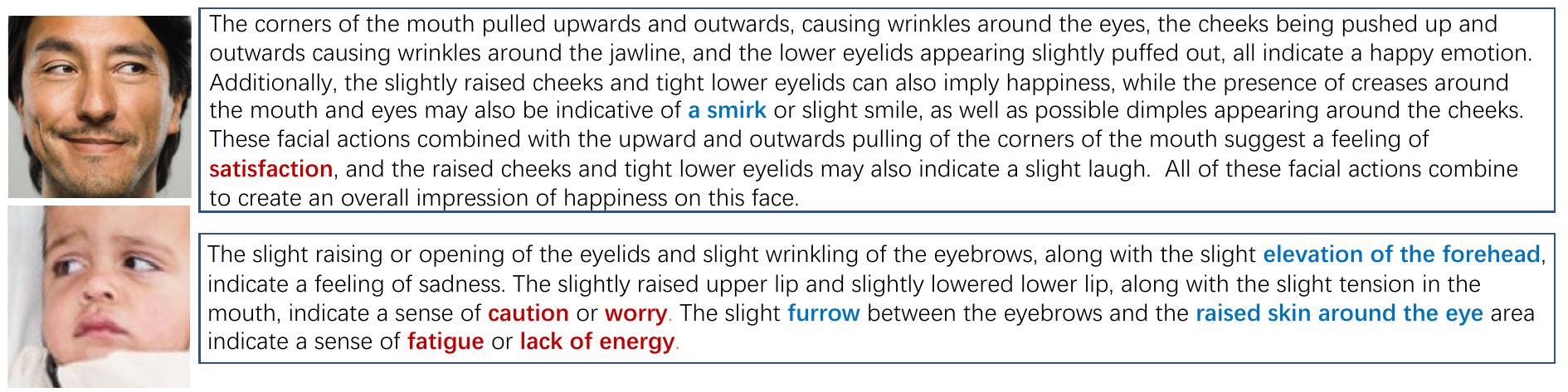}
    \vspace{-2 em}
    \caption{Examples of zero-shot ability of Exp-BLIP(ViT-G,OPT-6.7B)}
    \label{fig:zero-shot}
    \vspace{-1 em}
\end{figure}

\textbf{Zero-shot ability:} Fig.~\ref{fig:zero-shot} shows two examples of Exp-BLIP-generated descriptions of facial expressions. Exp-BLIP describes the unseen emotions (red) and facial actions (blue) within training data. The zero-shot ability is ascribed to the language model. 

\vspace{-1 em}

\vspace{-1 em}
\section{Conclusion}
\vspace{-0.4 em}
We introduce facial expression captioning as a novel task that aims to capture the nuanced facial actions and emotional states of a given facial image in natural language. To tackle this task, we propose a three-stage training framework that employs synthetic image-text pairs to train a facial expression captioner called Exp-BLIP. The experimental results showcase the powerful ability of Exp-BLIP and the intermediate visual representation. Nonetheless, Exp-BLIP suffers from certain limitations inherited from the pre-trained large models, such as generating repeated, syntax errors, or harmful content, which are inevitable in a frozen language model.
Exp-BLIP can be further enhanced by adopting more powerful LLMs and image encoders, and leveraging more data from other related tasks.

\section*{Acknowledgement}
This work is supported by National Natural Science Foundation of China (No. 62176248). We also thank ICT computing platform for providing GPUs. We thank Salesforce Research sharing the code of BLIP-2 via LAVIS~\cite{li2023lavis}. Our codes are based on LAVIS.

\bibliography{egbib}

\clearpage
\setcounter{section}{0}
\setcounter{figure}{0}
\setcounter{table}{0}
\setcounter{equation}{0}
\renewcommand{\thesection}{S\arabic{section}}
\renewcommand{\thefigure}{S\arabic{figure}}
\renewcommand{\thetable}{S\arabic{table}}
\renewcommand{\theequation}{S\arabic{equation}}
\renewcommand{\theHsection}{supp.\arabic{section}}
\renewcommand{\theHfigure}{supp.\arabic{figure}}
\renewcommand{\theHtable}{supp.\arabic{table}}
\renewcommand{\theHequation}{supp.\arabic{equation}}
\begin{center}
{\Large\bfseries Supplementary material to \emph{Describe Your Facial Expressions by Linking Image Encoders and Large Language Models}\par}
\end{center}

\vspace{-1 em}
\section{Descriptions of AU}
The training AU captions are synthesized in two ways: the rule-based text generator, and the GPT-based generator with one-shot prompt. For the rule-based AU captions labeling, the brief descriptions of each AU are listed in Table~\ref{tab:AU description} verb form. In order to ensure the syntactic correctness of synthesized captions, we appended s/es to the descriptions when merging them to the captions. If no AUs were labeled, we did not synthesize rule-based AU descriptions for the image. For GPT-based AU captions synthesizing, we first listed all 27 Action Units defined by the Facial Action Coding System (FACS), which were listed in Table~\ref{tab:AU description} noun form. Consequently, we conducted one-shot prompting to achieve GPT-based AU captions. If no AUs were labeled, we used a predetermined GPT-synthesized description in Sec.~\ref{pre} to avoid inaccurate outputs from GPT-3.5.

\vspace{-0.5 em}
\subsection{Brief descriptions of each AU}
Table~\ref{tab:AU description} shows the brief AU descriptions of noun form and verb form. The descriptions defined by FACS are of noun form. When synthesizing rule-based AU captions, the descriptions of verb form designed by us are more comprehensible for readers.

\begin{table}[!htbp]
    \caption{Brief AU descriptions of noun form and verb form.}
    \label{tab:AU description}
    \centering
    \footnotesize%
    \setlength{\tabcolsep}{6pt}%
    \renewcommand{\arraystretch}{1.2}%
    \begin{tabular}{l|c|c||l|c|c}
        \hline 
        AU  &   Noun form    &   Verb form & AU  &   Noun form    &   Verb form\\
        \hline
        AU1 & inner brow raiser & raise inner eyebrow & AU17 & chin raiser & raise chin\\
        AU2 & outer brow raiser & raise outer eyebrow & AU18 & lip pucker & pucker lips \\
        AU4 & brow lowerer & lower brow & AU19 & tongue show & stick tongue out \\
        AU5 & upper lid raiser & raise upper lid & AU20 & lip stretch & stretch lips \\
        AU6 & cheek raiser & raise cheek & AU21 & neck tightener & tighten neck \\
        AU7 & lid tightener & tighten lids & AU22 & lip funneler & show lip funneler  \\
        AU8 & lips toward each other & pull lips toward each other & AU23 & lip tightener & tighten lips  \\
        AU9 & nose wrinkler & wrinkle nose & AU24 & lip pressor & press lips \\
        AU10 & upper lip raiser & raise upper lip & AU25 & lips part & separate lips \\
        AU11 & nasolabial deepener & deepen nasolabial & AU26 & jaw drop & drop jaw  \\
        AU12 & lip corner puller & pull lip corner & AU27 & mouth stretch & stretch mouth \\
        AU14 & dimpler & tighten lip corner & AU28 & lip suck & suck lips \\
        AU15 & lip corner depressor & depress lip corner & AU43 & eyes closed & close eyes \\
        AU16 & lower lip depressor & depress lower lip & & & \\

        \hline
    \end{tabular}
\end{table}

\vspace{-0.5 em}
\subsection{Predetermined description of no AUs}\label{pre}

Due to the diversity of GPT-3.5's output and the high similarity of faces with no AUs labeled, we used a predetermined description which we selected from several outputs of GPT-3.5 for describing faces with no AUs. By conducting this strategy, we reduced the risk for GPT-3.5 to synthesize extra inaccurate descriptions for faces with no AUs. The predetermined description of no AUs is: \emph{
The eyes may be open and looking straight ahead, with the mouth closed or slightly open in a relaxed position. The forehead may be relatively smooth with minimal wrinkles or creases. The eyebrows may be in a natural position, not raised or furrowed. }

\section{Datasets}\label{dataset}
We used nearly 372k training image-text pairs in total. 
72k of the training data are derived from AU datasets BP4D~\cite{zhang2013high}, DISFA~\cite{mavadati2013disfa}, GFT~\cite{girard2017sayette}, RAF-AU~\cite{yan2020raf} and EmotioNet~\cite{fabian2016emotionet}.
Considering the high cost of using GPT-3.5 API\footnote{\url{https://platform.openai.com/docs/guides/gpt/chat-completions-api}} and the significant similarity among the consecutive frames in video-based AU datasets, we first select one sample from every ten frames. Then, we selected the frames which had different AU labels compared with the previous frame to fully utilize the datasets. This selection was conducted in training sets of video-based AU datasets (BP4D, DISFA, and GFT) and test set of GFT.
300k of the training data are derived from AffectNet~\cite{mollahosseini2017affectnet}, RAF-DB~\cite{li2017reliable} and FaceME~\cite{lu2019zero}.
The detail introduction for each dataset is in the following.

\textbf{BP4D} simultaneously records 2D and 3D facial expression videos in the lab. This dataset includes 41 participants (23 females and 18 males) with age ranging from 18 to 29 years. The released videos document the facial expression changes of the participants during eight different tasks. There are about 146,000 frames with 12 AU labels in the provided 2D videos. In our experiment, we split these frames into the training and test parts without overlapped subjects. In each video of training set, we selected one sample from every ten frames and also collected the samples which had different AU labels compared to the previous frame. Ultimately, there are 16627 frames of 28 subjects for training and 45805 frames of 13 subjects for testing. When evaluating the visual presentation in Tab.3 in the main text, we used the full training set.

\textbf{DISFA} collected spontaneous facial expressions of 27 participants while watching movie clips in the lab, annotating both the five intensity levels and 12 AU labels of each frame image. each video from the participant contains 4845 facial expression images and there are about 130,000 images in DISFA. %
We split these frames into the training and test parts without overlapped subjects. In each video of training set, we selected one sample from every ten frames and also collected the samples which had different AU labels compared to the previous frame. Ultimately, there are 14814 frames of 24 subjects for training and 14535 frames of 3 subjects for testing.

\textbf{GFT} records facial expression variations in 32 groups of three individuals during social gatherings, including 96 participants. It is the first dataset to capture facial expressions of multiple individuals in natural communication and interactive scenarios. The dataset we prepared comprises approximately 133,000 frames annotated with 10 AUs. In our experiment, we split these frames into the training and test parts without overlapped subjects. In each video, we selected one sample from every ten frames and also collected the samples which had different AU labels compared to the previous frame. Ultimately, there are 17719 frames of 78 subjects for training and 4034 frames of 18 subjects for testing.

\textbf{EmotioNet} contains over 1,000,000 facial expression data downloaded from the internet. It includes manual annotations for 50,000 images, specifying 11 AUs. The manually annotated images are divided into two parts: a validation set (publicly released) and a test set (not publicly released). Additionally, approximately 900,000 facial expression images were annotated using automated methods for AU and emotion labels, but these labels may contain noise. To ensure the accuracy of the data, we used the manual annotated images (approximately 21,000 images) labeled with 11 AUs for our experiment. We randomly split these images into the training and test parts. As a result, there are 19046 images in the training dataset and 2117 images in the test dataset.

\textbf{RAF-AU} contains 4,601 facial images annotated with 32 AUs in the wild. Some AU annotations distinguish the AU of upper, lower, left, and right regions of the face, by adding "T", "B", "L", "R" to the AU numbers, such as "17+B22+T24". In our experiment, we did not distinguish this different regions AU and removed the region signs, suggesting that the presence of AU in any region of the face was regarded as the presence of that AU. In addition, we only used the AUs shown in Table~\ref{tab:AU description} in this dataset. We randomly split all images into the training and test parts. As a result, there are 3733 images in the training dataset and 868 images in the test dataset.

\textbf{AffectNet} is a multi-class dataset collected from internet, including 287,618 training images annotated with 8 basic emotion classes: neutrality, happiness, sadness, anger, surprise, fear, disgust, contempt. Each AffectNet image only has one of the 8 emotion labels. In our experiment, we used all the official training and test images. The test dataset comprises 4000 images, with each of the 8 basic classes containing 500 images.

\textbf{RAF-DB} is a multi-class dataset containing of a 7-class basic emotion part and a 12-class compound emotion part. In our experiments, we used all the compound label part, where the labels were formed by combining 2 of the 6 classes: happy, sad, anger, surprise, fear, disgust. We randomly split all the 3954 images into the training and test parts. As a result, there are 3162 images for training and 792 images for testing. %

\textbf{FaceME} is a multi-label facial expression dataset collected from the internet, containing 10,062 images and 85 labels. The labels not only include emotions, but also labels about action, health, and inward thoughts. Each image is labelled by 3 annotators. To enhance the credibility of the labels, we only used the labels that are annotated presence by at least 2 annotators. In our experiment, we used all images for training without dividing test dataset.

\section{Visual representations}
The first part of this section re-conducted the visual representations experiment introduced in Sec. 4.2 by combing all the five AU training sets and further conducted this experiment with a fine-tuning strategy. The second part discusses the comparison with SOTA AU detector (DGCN\cite{jia2023novel}) measured by $Acc.\%$ on RAF-AU.

\begin{table}[!htbp]
    \caption{Performance on BP4D and AffectNet. (ViT-B/OPT-2.7B)}
    \label{tab:BP4D vitb}
    \centering
    \scriptsize%
    \setlength{\tabcolsep}{2pt}%
    \renewcommand{\arraystretch}{1.2}%
    \begin{tabular}{lccccccccccccc|c}
        \hline
        \multirow{2}{*}{Models} & \multicolumn{13}{c|}{BP4D (F1 score$\times 100$)} & AffectNet (\textit{Acc.}\%)\\
        & AU1& AU2& AU4& AU6& AU7& AU10& AU12& AU14& AU15& AU17& AU23& AU24 & Avg & Emotion\\
        \hline
        ViT-B              & 46.3 & 38.2 & 50.8 & 79.3 & 73.4 & 85.4 & 87.9 & \textbf{69.3} & 24.0 & 53.0 & 0.9 & 24.2 & 52.7 & 37.8 \\
        ViT-B$^\text{fine-tune}$ & 55.1 & \textbf{65.3} & \underline{64.8} & 53.3 & 59.4 & 83.9 & 88.4 & 5.9 & 43.5 & 22.2 & \textbf{96.0} & \textbf{87.3} & \underline{60.4} & 51.4\\
        \hline
        ViT-B(Emot)  & 47.0 & 37.5 & 52.9 & 79.0 & 73.0 & 85.8 & 88.6 & 68.0 & 24.3 & 54.3 & 7.7 & 21.9 & 53.3 & 38.2 \\
        QFormer(Emot)  & 52.1 & 44.7 & \textbf{66.7} & 80.8 & 74.8 & 85.2 & 88.6 & 50.4 & 36.9 & 59.2 & 14.3 & 28.5 & 56.9 & \textbf{54.3} \\
        ViT-B(AU)    & 47.3 & 39.6 & 59.6 & 78.9 & 73.3 & 85.7 & 88.6 & 68.2 & 23.3 & 52.7 & 2.1 & 15.8 & 52.9 & 38.8 \\
        QFormer(AU)    & \textbf{56.4} & 45.6 & 57.6 & 80.3 & 74.3 & 86.3 & \underline{89.9} & 61.8 & 41.1 & \textbf{61.7} & \underline{33.2} & \underline{42.1} & \textbf{60.9} & 45.2 \\
        ViT-B(Mix)   & 47.6 & 39.0 & 51.4 & 79.7 & 73.2 & \textbf{87.1} & 89.4 & 67.1 & 21.2 & 53.1 & 3.0 & 20.1 & 52.7  & 38.6 \\
        QFormer(Mix)   & 53.0 & \underline{50.1} & 52.4 & \underline{81.0} & 74.8 & \textbf{87.1} & \textbf{90.3} & \underline{68.6} & \textbf{49.8} & 56.6 & 25.0 & 35.0 & 60.3 & 52.1 \\
        ViT-B(Cat)  & 48.5 & 36.6 & 55.6 & 78.2 & 73.3 & 86.8 & 88.6 & 64.2 & 24.9 & 52.9 & 6.2 & 16.6 & 52.7  & 39.8 \\
        QFormer(Cat)  & 54.2 & 43.7 & 56.7 & \textbf{81.1} & \textbf{75.9} & 87.0 & 88.9 & 66.2 & \underline{47.5} & 56.2 & 19.6 & 22.2 & 58.3 & 48.1 \\
        ViT-B(Exp)  & 46.3 & 38.0 & 57.0 & 78.8 & 72.4 & 86.0 & 88.3 & 65.4 & 26.3 & 53.8 & 7.2 & 18.8 & 53.2  & 39.6 \\
        QFormer(Exp)  & \underline{55.7} & 44.8 & 58.3 & 80.5 & \underline{75.1} & 86.9 & 89.6 & 67.4 & 43.7 & 56.4 & 24.0 & 41.1 & 60.3  & 48.2 \\
        \hline
    \end{tabular}
    \vspace{-1 em}
\end{table}

\begin{table}[!htbp]
    \caption{Performance on RAF-AU. (ViT-B/OPT-2.7B)}
    \label{tab:RAFAU vitb}
    \centering
    \scriptsize%
    \setlength{\tabcolsep}{1pt}%
    \renewcommand{\arraystretch}{1.2}%
    \begin{tabular}{lcccccccccccccccccc}
        \hline
        Models & AU1& AU2& AU4& AU5& AU6& AU7& AU9& AU10& AU12& AU15& AU16& AU17& AU22 &  AU24 & AU25& AU26& AU27& Avg    \\
        \hline
        ViT-B             & 48.4 & 49.9 & 67.9 & 23.8 & 32.4 & 19.7 & 25.1 & 47.5 & 69.1 & 15.5 & 24.5 & 18.4 & 2.0 & 16.9 & 86.7 & 32.3 & 58.9 & 37.6   \\
        ViT-B$^\text{fine-tune}$ & 66.3 & 53.9 & \textbf{79.5} & \underline{67.1} & 36.7 & \textbf{32.5} & \textbf{79.6} & \textbf{72.9} & 66.4 & \textbf{46.8} & \textbf{57.7} & \textbf{68.0} & \textbf{19.6} & \textbf{26.5} & \underline{94.7} & \textbf{67.7} & \textbf{81.8} & \textbf{59.9}\\
        \hline
        ViT-B(Emot)   & 48.1 & 50.5 & 68.9 & 27.2 & 34.8 & 23.0 & 28.0 & 50.2 & 71.1 & 16.4 & 27.6 & 19.8 & 2.2 & 18.7 & 87.4 & 29.2 & 65.3 & 39.3 \\
        QFormer(Emot) & 54.2 & 53.9 & 72.8 & 58.3 & 35.7 & \underline{26.7} & 53.4 & 62.8 & 62.4 & 26.1 & 49.5 & 49.9 & \underline{5.3} & 11.7 & 92.5 & 42.1 & \underline{76.6} & 49.1 \\
        ViT-B(AU)    & 50.3 & 52.6 & 71.0 & 30.8 & 29.6 & 22.2 & 35.5 & 52.4 & 70.8 & 16.1 & 29.8 & 28.1 & 4.4 & \underline{19.2} & 87.6 & 36.2 & 67.1 & 41.4 \\
        QFormer(AU)    & \underline{66.6} & \underline{66.1} & 78.0 & 63.3 & 31.5 & 25.0 & 69.7 & 54.5 & 61.5 & 19.6 & 39.6 & \underline{62.5} & 3.6 & 9.7 & \underline{94.7} & \underline{55.3} & 69.6 & 51.2 \\
        ViT-B(Mix)   & 47.9 & 53.3 & 68.5 & 32.0 & 34.4 & 25.9 & 29.3 & 51.6 & 69.7 & 15.9 & 27.1 & 27.4 & 2.4 & 18.3 & 88.7 & 33.6 & 65.4 & 40.7 \\
        QFormer(Mix)   & \textbf{67.9} & \textbf{68.4} & \underline{79.2} & 62.5 & \textbf{39.1} & 18.0 & 77.2 & \underline{65.9} & \underline{72.3} & \underline{33.6} & \underline{54.0} & 45.9 & 0.0 & 8.7 & 92.5 & 53.9 & \underline{76.6} & \underline{53.9} \\
        ViT-B(Cat)   & 49.5 & 52.0 & 69.1 & 33.8 & 33.2 & 22.4 & 35.1 & 48.8 & 71.1 & 16.5 & 22.2 & 26.5 & 2.5 & 18.0 & 88.8 & 31.6 & 65.8 & 40.4 \\
        QFormer(Cat)  & 66.1 & 65.3 & 77.2 & 60.9 & 30.3 & 8.4 & \underline{78.2} & 63.6 & 59.6 & 25.7 & 38.8 & 60.6 & 0.0 & 12.6 & \underline{94.7} & 51.9 & 71.8 & 50.9 \\
        ViT-B(Exp)   & 48.7 & 52.7 & 68.6 & 30.2 & 32.9 & 25.9 & 32.0 & 52.8 & 70.8 & 16.1 & 23.5 & 23.2 & 4.6 & \underline{19.2} & 88.4 & 35.4 & 66.4 & 40.7 \\
        QFormer(Exp)  & 66.2 & 63.6 & 76.1 & \textbf{68.2} & \underline{38.2} & 12.2 & 74.3 & 60.2 & \textbf{72.7} & 22.3 & 38.2 & 54.9 & 2.2 & 9.1 & \textbf{95.5} & 46.1 & 65.4 & 50.9 \\
        \hline
    \end{tabular}
    \vspace{-1 em}
\end{table}

\begin{table}[!htbp]
    \caption{Performance on BP4D and AffectNet. (ViT-G/OPT-6.7B)}
    \label{tab:BP4D vitg}
    \centering
    \scriptsize%
    \setlength{\tabcolsep}{2pt}%
    \renewcommand{\arraystretch}{1.2}%
    \begin{tabular}{lccccccccccccc|c}
        \hline
        \multirow{2}{*}{Models} & \multicolumn{13}{c|}{BP4D (F1 score$\times 100$)} & AffectNet (\textit{Acc.}\%)\\
        & AU1& AU2& AU4& AU6& AU7& AU10& AU12& AU14& AU15& AU17& AU23& AU24 & Avg & Emotion\\
        \hline
        ViT-G              & 34.2 & 16.7 & 48.6 & \textbf{80.1} & 70.0 & 85.9 & 89.6 & 59.6 & 15.7 & 51.8 & 19.1 & 17.1  & 49.0 & 44.8\\
        ViT-G$^\text{fine-tune}$ & 44.5 & 30.6 & 57.4 & 79.4 & 69.2 & 86.0 & 88.8 & 62.1 & 39.2 & 59.9 & \textbf{41.7} & 30.9  & 57.5 & 50.3\\
        \hline
        ViT-G(Emot)  & 35.5 & 16.2 & 53.9 & 78.1 & 73.8 & \textbf{87.5} & 89.7 & \textbf{70.5} & 35.0 & 56.7 & 33.0 & 21.1 & 54.3 & 50.9\\
        QFormer(Emot)  & 5.9 & 3.8 & 43.6 & 71.7 & 74.4 & 82.9 & 87.7 & 30.0 & 4.3 & 9.8 & 0.1 & 12.4 & 35.6 & \textbf{52.7} \\
        ViT-G(AU)    & 56.3 & 49.7 & \underline{66.2} & 79.5 & 75.3 & 85.9 & 89.0 & 59.9 & \underline{46.0} & \textbf{70.4} & \underline{41.2} & 40.0  & \underline{63.3} & 50.1\\
        QFormer(AU)    & \underline{58.2} & \underline{50.6} & \textbf{67.2} & 79.3 & 75.8 & \underline{86.6} & \textbf{89.8} & 63.9 & \textbf{50.8} & \underline{65.1} & 36.5 & 42.4  & \textbf{63.9} & 52.2 \\
        ViT-G(Mix)   & 55.1 & 41.2 & 56.4 & 76.9 & 73.6 & 83.2 & 84.5 & 57.7 & 43.7 & 59.0 & 34.3 & 36.4 & 58.5  & 48.2 \\
        QFormer(Mix)   & 54.5 & 43.8 & 56.2 & 79.9 & 75.7 & 84.8 & 89.5 & \underline{68.3} & 44.9 & 57.2 & 37.4 & 33.8  &60.5 & 49.3 \\
        ViT-G(Cat)  & 52.5 & 40.1 & 55.0 & 73.6 & 73.8 & 82.4 & 87.7 & 62.9 & 45.3 & 58.9 & 30.8 & \underline{43.3} & 58.9  & 48.9 \\
        QFormer(Cat)  & 54.9 & 44.4 & 49.8 & \textbf{80.1} & 76.0 & 85.8 & \textbf{89.8} & 67.2 & 42.9 & 58.0 & 33.2 & 36.0 & 59.8  & 49.5 \\
        ViT-G(Exp)  & \textbf{59.0} & 50.2 & 59.5 & 78.5 & \underline{77.7} & 86.5 & 89.4 & 64.3 & 36.0 & 62.5 & 39.9 & \textbf{55.6} & \underline{63.3}  & 51.0 \\
        QFormer(Exp)  & 57.5 & \textbf{54.1} & 58.7 & 79.4 & \textbf{78.1} & 86.5 & 89.3 & 64.2 & 39.9 & 62.9 & 32.1 & 40.5 & 61.9  & 51.6 \\
        \hline
    \end{tabular}
    \vspace{-1 em}
\end{table}

\begin{table}[!htbp]
    \caption{Performance on RAF-AU. (ViT-G/OPT-6.7B)}
    \label{tab:RAFAU vitg}
    \centering
    \scriptsize%
    \setlength{\tabcolsep}{1pt}%
    \renewcommand{\arraystretch}{1.2}%
    \begin{tabular}{lcccccccccccccccccc}
        \hline
        Models & AU1& AU2& AU4& AU5& AU6& AU7& AU9& AU10& AU12& AU15& AU16& AU17& AU22 &  AU24 & AU25& AU26& AU27& Avg    \\
        \hline
        ViT-G             & 48.9 & 50.0 & 71.0 & 65.4 & 26.1 & 12.1 & 64.5 & 53.1 & 60.7 & 9.8 & 50.0 & 48.9 & 10.4 & 9.7 & 90.0 & 28.4 & 75.1 & 45.5  \\
        ViT-G$^\text{fine-tune}$ & 67.4 & 64.5 & 78.7 & 64.5 & 34.6 & \underline{29.5} & 67.2 & \textbf{68.7} & \textbf{67.2} & 45.8 & \textbf{61.2} & 52.4 & 25.2 & \underline{23.7} & 94.2 & \textbf{65.6} & \textbf{81.5}  & \underline{58.3}\\
        \hline
        ViT-G(Emot)   & 49.8 & 29.9 & 74.7 & 66.5 & 14.9 & 9.9 & 59.3 & 59.4 & 58.7 & 28.9 & 51.8 & 47.6 & 5.3 & 14.0 & 88.7 & 41.0 & 74.2 & 45.6\\
        QFormer(Emot) & 30.4 & 9.6 & 73.1 & 58.9 & 8.4 & 16.5 & 47.1 & 47.0 & 49.9 & 0.0 & 34.0 & 5.0 & 0.0 & 0.0 & 86.9 & 16.1 & 73.4 & 32.7\\
        ViT-G(AU)    & 72.4 & 65.4 & 80.7 & \underline{75.9} & \textbf{47.8} & 27.3 & \textbf{76.2} & 60.3 & 63.2 & 35.4 & 53.1 & 55.4 & \textbf{30.4} & 15.9 & 93.6 & 58.1 & 77.3  & 58.1\\
        QFormer(AU)    & 74.6 & \textbf{72.1} & 81.5 & 71.2 & \underline{46.5} & \textbf{40.5} & 75.3 & \underline{64.0} & \underline{66.8} & 45.7 & 46.7 & \textbf{69.0} & \underline{28.3} & \textbf{32.4} & 94.3 & \underline{61.5} & 78.3  & \textbf{61.7}\\
        ViT-G(Mix)   & 76.1 & 66.7 & 83.6 & \textbf{77.3} & 29.8 & 26.7 & 73.4 & 49.1 & 57.6 & \textbf{48.2} & 36.8 & 36.1 & 2.8 & 10.5 & 94.9 & 38.1 & 74.7 & 51.9\\
        QFormer(Mix)   & \underline{79.9} & 66.4 & \underline{84.4} & 74.4 & 25.9 & 24.3 & 75.5 & 56.2 & 62.9 & 36.5 & 37.4 & 54.5 & 0.0 & 3.5 & 94.6 & 32.4 & \underline{79.2} & 52.2\\
        ViT-G(Cat)   & 77.4 & 66.4 & \textbf{85.3} & 75.2 & 28.0 & 23.7 & 75.0 & 49.1 & 65.2 & \underline{46.0} & 30.8 & 51.7 & 2.8 & 3.4 & \underline{95.3} & 43.6 & 74.8 & 52.6\\
        QFormer(Cat)  & \textbf{82.0} & \underline{71.3} & \underline{84.4} & 72.8 & 24.6 & 21.4 & 74.7 & 54.9 & 63.7 & 38.1 & 40.9 & 58.0 & 0.0 & 11.4 & \textbf{95.4} & 36.1 & 74.2 & 53.2\\
        ViT-G(Exp)   & 70.6 & 66.9 & 80.1 & 71.9 & 40.9 & 21.4 & \underline{75.8} & 59.5 & 58.2 & 45.5 & \underline{57.8} & \underline{58.8} & 22.5 & 9.8 & 93.4 & 60.4 & 74.7 & 57.0\\
        QFormer(Exp)  & 71.8 & 70.9 & 81.8 & 70.6 & 32.8 & 25.6 & 75.0 & 63.6 & 58.7 & 26.5 & 49.5 & 56.5 & 22.7 & 7.0 & 93.5 & 56.8 & 73.2 & 55.1\\
        \hline
    \end{tabular}
    \vspace{-1 em}
\end{table}

\begin{table}[!htbp]
    \caption{Performance of fine-tuning on BP4D and AffectNet. (ViT-B/OPT-2.7B)}
    \label{tab:BP4D vitb ft}
    \centering
    \scriptsize%
    \setlength{\tabcolsep}{1.5pt}%
    \renewcommand{\arraystretch}{1.2}%
    \begin{tabular}{lccccccccccccc|c}
        \hline
        \multirow{2}{*}{Models} & \multicolumn{13}{c|}{BP4D (F1 score$\times 100$)} & AffectNet (\textit{Acc.}\%)\\
        & AU1& AU2& AU4& AU6& AU7& AU10& AU12& AU14& AU15& AU17& AU23& AU24 & Avg & Emotion\\
        \hline
        ViT-B$^\text{fine-tune}$ & 55.1 & \textbf{65.3} & 64.8 & 53.3 & 59.4 & 83.9 & 88.4 & 5.9 & 43.5 & 22.2 & \textbf{96.0} & \textbf{87.3} & 60.4 & 51.4\\
        \hline
        ViT-B(Emot)$^\text{fine-tune}$  & 55.2 & 44.0 & 60.0 & 78.4 & 73.2 & 85.6 & 89.4 & 62.0 & 44.9 & 61.5 & 40.2 & 44.1 & 61.5 & 52.9 \\
        QFormer(Emot)$^\text{fine-tune}$  & 55.1 & 47.5 & 62.4 & 77.8 & 71.0 & \textbf{86.1} & 89.5 & 60.2 & 47.2 & 61.6 & 43.9 & 47.4 & 62.5 & 54.7 \\
        ViT-B(AU)$^\text{fine-tune}$    & 55.6 & 44.4 & 62.5 & 77.8 & 72.6 & 85.5 & 89.3 & 58.3 & 45.5 & 61.6 & 39.2 & \underline{47.8} & 61.7 & 51.9 \\
        QFormer(AU)$^\text{fine-tune}$    & 54.5 & 48.8 & 62.4 & \underline{80.1} & \underline{75.2} & 85.8 & 90.0 & \textbf{66.9} & 49.6 & \textbf{64.5} & 46.5 & 45.0 & \underline{64.1} & 53.6 \\
        ViT-B(Mix)$^\text{fine-tune}$   & \underline{55.8} & 43.6 & 59.2 & 78.5 & 73.8 & \textbf{86.1} & 89.5 & \underline{64.7} & 43.6 & 60.6 & 39.9 & 47.0 & 61.9  & 52.6 \\
        QFormer(Mix)$^\text{fine-tune}$   & \textbf{56.9} & 50.9 & \underline{65.1} & 78.6 & 74.1 & 85.3 & \textbf{90.9} & 60.5 & \underline{49.8} & 62.6 & 44.8 & 47.2 & 63.9 & 54.3 \\
        ViT-B(Cat)$^\text{fine-tune}$  & 54.1 & 45.8 & 61.8 & 78.3 & 72.9 & 85.4 & 89.1 & 62.0 & 44.4 & 60.4 & 42.5 & 46.2 & 61.9  & 52.1 \\
        QFormer(Cat)$^\text{fine-tune}$  & 54.6 & 47.3 & 63.4 & 78.8 & 74.0 & 85.4 & 89.7 & 63.0 & 48.9 & 64.1 & \underline{46.8} & 47.2 & 63.6 & 54.3\\
        ViT-B(Exp)$^\text{fine-tune}$  & 55.6 & 43.4 & 60.0 & 77.8 & 73.2 & 86.0 & 89.9 & 61.9 & 44.1 & 61.6 & 42.5 & 41.7 & 61.5  & 52.1 \\
        QFormer(Exp)$^\text{fine-tune}$  & \underline{55.8} & \underline{51.3} & \textbf{67.0} & \textbf{80.2} & \textbf{75.5} & 85.9 & \underline{90.1} & 62.1 & \textbf{50.3} & \textbf{64.5} & 45.7 & 43.8 & \textbf{64.4}  & \textbf{54.8} \\
        \hline
    \end{tabular}
\end{table}

\begin{table}[!htbp]
    \caption{Performance of fine-tuning on RAF-AU. (ViT-B/OPT-2.7B)}
    \label{tab:RAFAU vitb ft}
    \centering
    \scriptsize%
    \setlength{\tabcolsep}{0.5pt}%
    \renewcommand{\arraystretch}{1.2}%
    \begin{tabular}{lcccccccccccccccccc}
        \hline
        Models & AU1& AU2& AU4& AU5& AU6& AU7& AU9& AU10& AU12& AU15& AU16& AU17& AU22 &  AU24 & AU25& AU26& AU27& Avg    \\
        \hline
        ViT-B$^\text{fine-tune}$ & 66.3 & 53.9 & 79.5 & 67.1 & 36.7 & 32.5 & 79.6 & 72.9 & 66.4 & 46.8 & 57.7 & 68.0 & 19.6 & 26.5 & 94.7 & 67.7 & 81.8 & 59.9\\
        \hline
        ViT-B(Emot)$^\text{fine-tune}$   & 68.5 & 67.3 & 81.6 & 62.5 & 28.2 & 27.6 & 76.6 & 69.9 & 66.3 & 49.1 & 51.1 & 64.9 & 9.6 & 26.8 & 95.4 & 56.2 & 78.4 & 57.6\\
        QFormer(Emot)$^\text{fine-tune}$ & 75.0 & 71.4 & 83.6 & 74.0 & 46.6 & 31.2 & \textbf{81.8} & 71.9 & \underline{74.0} & 60.3 & 65.7 & 72.9 & 19.8 & 23.5 & 95.8 & 72.9 & \textbf{86.5} & 65.1\\
        ViT-B(AU)$^\text{fine-tune}$    & 69.3 & 66.7 & 82.3 & 67.5 & 28.3 & 32.2 & 75.5 & 71.3 & 64.6 & 49.1 & 46.8 & 62.4 & 18.2 & 20.9 & 95.3 & 54.6 & 77.5 & 57.8\\
        QFormer(AU)$^\text{fine-tune}$    & 73.9 & 69.1 & \textbf{83.9} & 74.5 & \underline{48.9} & \underline{36.8} & \underline{81.3} & \textbf{77.2} & 70.8 & \underline{63.5} & 66.4 & \underline{74.9} & 26.2 & \textbf{39.0} & 95.7 & \underline{74.7} & 84.8 & 67.2\\
        ViT-B(Mix)$^\text{fine-tune}$   & 70.4 & 66.9 & 82.9 & 68.4 & 23.6 & 28.6 & 77.2 & 71.3 & 65.0 & 48.1 & 45.8 & 63.5 & 17.6 & 34.2 & 95.0 & 49.8 & 75.2 & 57.9\\
        QFormer(Mix)$^\text{fine-tune}$   & \textbf{79.1} & \textbf{80.4} & \textbf{83.9} & \textbf{77.6} & \textbf{51.5} & 30.8 & 81.1 & 74.9 & 73.2 & \textbf{68.8} & \underline{69.4} & \textbf{75.0} & \textbf{34.3} & 37.6 & \textbf{96.6} & \textbf{77.8} & \underline{85.3} & \textbf{69.3}\\
        ViT-B(Cat)$^\text{fine-tune}$   & 71.2 & 67.4 & 81.6 & 67.8 & 25.8 & 22.2 & 77.0 & 70.1 & 66.7 & 54.5 & 49.3 & 66.3 & 18.4 & 34.8 & 95.0 & 54.8 & 75.7 & 58.7\\
        QFormer(Cat)$^\text{fine-tune}$  & \underline{75.2} & 70.2 & 82.5 & 75.1 & 42.3 & 30.4 & 79.6 & \underline{76.4} & 71.2 & 62.5 & \textbf{70.5} & 74.0 & \underline{27.4} & \underline{37.8} & 95.2 & 73.1 & 84.3 & 66.4\\
        ViT-B(Exp)$^\text{fine-tune}$   & 70.4 & 66.9 & 82.8 & 65.9 & 26.8 & 33.9 & 76.9 & 68.5 & 64.3 & 48.1 & 44.6 & 62.2 & 20.2 & 28.0 & 95.0 & 50.0 & 74.2 & 57.6\\
        QFormer(Exp)$^\text{fine-tune}$  & 71.7 & \underline{72.3} & 82.2 & \underline{77.0} & 47.5 & \textbf{44.1} & 80.1 & 74.3 & \textbf{75.1} & 62.1 & 68.7 & 74.8 & 25.3 & 35.5 & \underline{96.1} & 74.1 & 85.0 & \underline{67.4}\\
        \hline
    \end{tabular}
\end{table}

\subsection{Combing AU training sets}
Tables \ref{tab:BP4D vitb} and \ref{tab:RAFAU vitb} report the performance of visual representations, including F1 scores on BP4D and RAF-AU datasets, and classification accuracy on AffectNet dataset. Different from the experiment in Sec. 4.2, these classifiers were trained on the combination of full EmotioNet, RAFAU training sets and sampled BP4D, DISFA and GFT training sets, rather than single datasets. The sampling strategy is introduced in Sec.~\ref{dataset}. 
In addition, we selected more AUs in RAF-AU for the experiment.
ViT-B stands for the pre-trained image encoder. The values are reported under a linear probe strategy.
ViT-B$^\text{fine-tune}$ is ViT-B fine-tuned on the training data for AU detection or emotion classification, using a fine-tuning strategy.
Other models with the names formatted as <ViT-B/QFormer>(AU/Emot/Mix/Cat/Exp) denote the visual representations (e.g., the image encoder or Q-Former) from the varied models (i.e., AU-BLIP, Emot-BLIP, Mix-BLIP, Cat-BLIP, Exp-BLIP).
For these representations, linear probe strategies were applied. The language model trained with ViT-B/QFormer is OPT2.7B.
All the models were trained for 20 epochs except for ViT-B$^\text{fine-tune}$ which was trained for 50 epochs. Tables \ref{tab:BP4D vitg} and \ref{tab:RAFAU vitg} report the performance of classifiers trained following the same settings above, except for the use of image encoder ViT-G and language model OPT-6.7B.

Tables \ref{tab:BP4D vitb} and \ref{tab:RAFAU vitb} show that all of the ViT-B and Q-Former models with linear probe strategy in our approach outperform the ViT-B baseline. This suggests that incorporating language tasks improves the visual representations and enhances performance on related downstream tasks. It is also observed that the features of Q-Former are superior to those of the corresponding image encoder, suggesting a stronger visual representation of Q-Former than the image encoder. This observation is consistent with the main text and Tables \ref{tab:BP4D vitg} \& \ref{tab:RAFAU vitg}.

Compared with ViT-B baseline with linear probe strategy, ViT-B$^\text{fine-tune}$ with fine-tuning strategy achieves better performance on both AU detection and emotion classification tasks. To explore the advantages of fine-tuning strategy, we fine-tuned each model in Tables \ref{tab:BP4D vitb} and \ref{tab:RAFAU vitb} and the results are shown in Tables \ref{tab:BP4D vitb ft} and \ref{tab:RAFAU vitb ft}. The settings of the models in Tables \ref{tab:BP4D vitb ft} and \ref{tab:RAFAU vitb ft} are the same as those in Tables \ref{tab:BP4D vitb} and \ref{tab:RAFAU vitb} except for the fine-tuning strategy.

It is observed that most of the ViT-B and Q-Former models with fine-tuning strategy in our approach perform better than the ViT-B$^\text{fine-tune}$ baseline, further indicating the superiority of our approach in visual representation. Comparing the linear probe results in Tab. \ref{tab:BP4D vitb} and \ref{tab:RAFAU vitb} with the corresponding fine-tuning results in Tab. \ref{tab:BP4D vitb ft} and \ref{tab:RAFAU vitb ft}, all the fine-tuning results show an improvement over the linear probe results, indicating that fine-tuning is a more effective strategy for extracting excellent features when the diversity of training data is large.

\begin{table}[t]
    \caption{Performance of visual representation on RAFAU. ($Acc.\%$; $^*$: original values.)}
    \label{tab:RAFAU vitg acc}
    \centering
    \scriptsize%
    \setlength{\tabcolsep}{3.7pt}%
    \renewcommand{\arraystretch}{1.2}%
    \begin{tabular}{lcccccccccccccc}
        \hline
        Models & AU1& AU2& AU4& AU5& AU6& AU7& AU9& AU12& AU17 & AU23& AU24& AU25& \textbf{Avg} & Avg(F1)    \\
        \hline
        FaRL\cite{zheng2022general}         & 69.0  & 71.0  & 76.6  & 70.6  & 61.6  & 68.8  & 69.2  & 75.0  & 70.3  & 66.7  & 81.1  & 69.2  & 70.8 & 55.7\\
        DGCN$^*$\cite{jia2023novel}      & 79.8 & 87.1 & 68.5 & 85.5 & \textbf{91.9} & \textbf{93.4} & 81.1 & 79.8 & \textbf{86.5} & \textbf{98.4} & \textbf{96.6} & 65.9 & \underline{84.5}  & - \\
        \hline
        ViT-G             & 72.7 & 76.2 & 80.5 & 76.6 & 69.5 & 70.6 & 77.6 & 81.2 & 74.4 & 76.4 & 71.2 & 89.3 & 76.3  & 53.5     \\
        ViT-G$^\text{fine-tune}$ & 83.5 & \textbf{89.3} & 80.9 & \underline{87.3} & \underline{89.7} & \underline{82.3} & 75.7 & \underline{84.6} & 73.5 & 85.5 & 36.1 & 58.3 & 77.2  &  45.2  \\
        \hline
        ViT-G(Emot)   & 73.4 & 72.5 & 82.3 & 77.6 & 71.0 & 65.2 & 81.3 & 80.6 & 76.6 & 75.0 & 68.9 & 87.0 & 76.0   & 53.0  \\
        QFormer(Emot) & 70.6 & 74.8 & 80.8 & 80.1 & 73.0 & 73.0 & 81.5 & 77.1 & 73.5 & 84.9 & 72.6 & 85.1 & 77.2   & 52.2 \\
        ViT-G(AU)    & 85.5 & 84.1 & 85.1 & \textbf{87.8} & 85.5 & 72.4 & \textbf{91.9} & 83.5 & 83.9 & 76.8 & 76.2 & 90.1 & 83.6   & \underline{59.9}  \\
        QFormer(AU)  & \textbf{87.2} & 87.4 & 84.1 & 84.1 & 82.9 & 70.7 & \underline{90.2} & 84.0 & \underline{85.8} & 81.9 & \underline{82.6} & 94.7 & \textbf{84.6}  & \textbf{60.2}  \\
        ViT-G(Mix)   & 77.6 & 78.0 & 84.8 & 80.4 & 78.1 & 72.5 & 85.0 & 82.3 & 74.0 & 93.8 & 74.2 & 94.5 & 81.3   & 57.4 \\
        QFormer(Mix)   & 82.8 & 80.3 & \underline{85.1} & 82.8 & 83.3 & 75.8 & 87.3 & 83.3 & 80.2 & \underline{94.0} & 77.5 & 94.5 & 83.9   & 59.4  \\
        ViT-G(Cat)   & 80.9 & 79.8 & 84.9 & 81.7 & 77.8 & 73.7 & 85.0 & 82.0 & 75.1 & 92.3 & 74.2 & \underline{95.2} & 81.9  & 57.2  \\
        QFormer(Cat)  & \underline{86.3} & 84.2 & \textbf{85.4} & 85.1 & 81.5 & 73.3 & 86.4 & \underline{84.6} & 83.1 & 85.6 & 79.5 & \textbf{95.9} & 84.2   & 59.5\\
        ViT-G(Exp)   & 82.4 & 83.3 & 84.4 & 85.6 & 76.8 & 69.4 & 86.2 & 84.4 & 80.8 & 72.8 & 72.1 & 91.5 & 80.8   & 57.5 \\     
        QFormer(Exp) & 85.7 & \underline{87.8} & 85.3 & 83.9 & 79.6 & 71.8 & 89.2 & \textbf{84.8} & 79.3 & 77.0 & 76.3 & 93.3 & 82.8   & 59.7  \\
        \hline
    \end{tabular}
\end{table}

\subsection{Comparison with DGCN measured by Accuracy}
Table \ref{tab:RAFAU vitg acc} reports the classification accuracy on RAF-AU dataset, including the comparison with the SOTA AU detector (DGCN\cite{jia2023novel}). It is observed that the performance of QFormer(AU) is compatible with that of DGCN. However, classification accuracy is not a prevalent metric for AU detection evaluation, because of the unbalanced presence of each AU. Consequently, we add the Average F1 score of each method to Table \ref{tab:RAFAU vitg acc}. The observation of performance evaluated by F1 score is consistent with those in the main text.

\vspace{-1 em}
\begin{figure}[!htbp]
    \centering
    \includegraphics[width=1.0\textwidth]{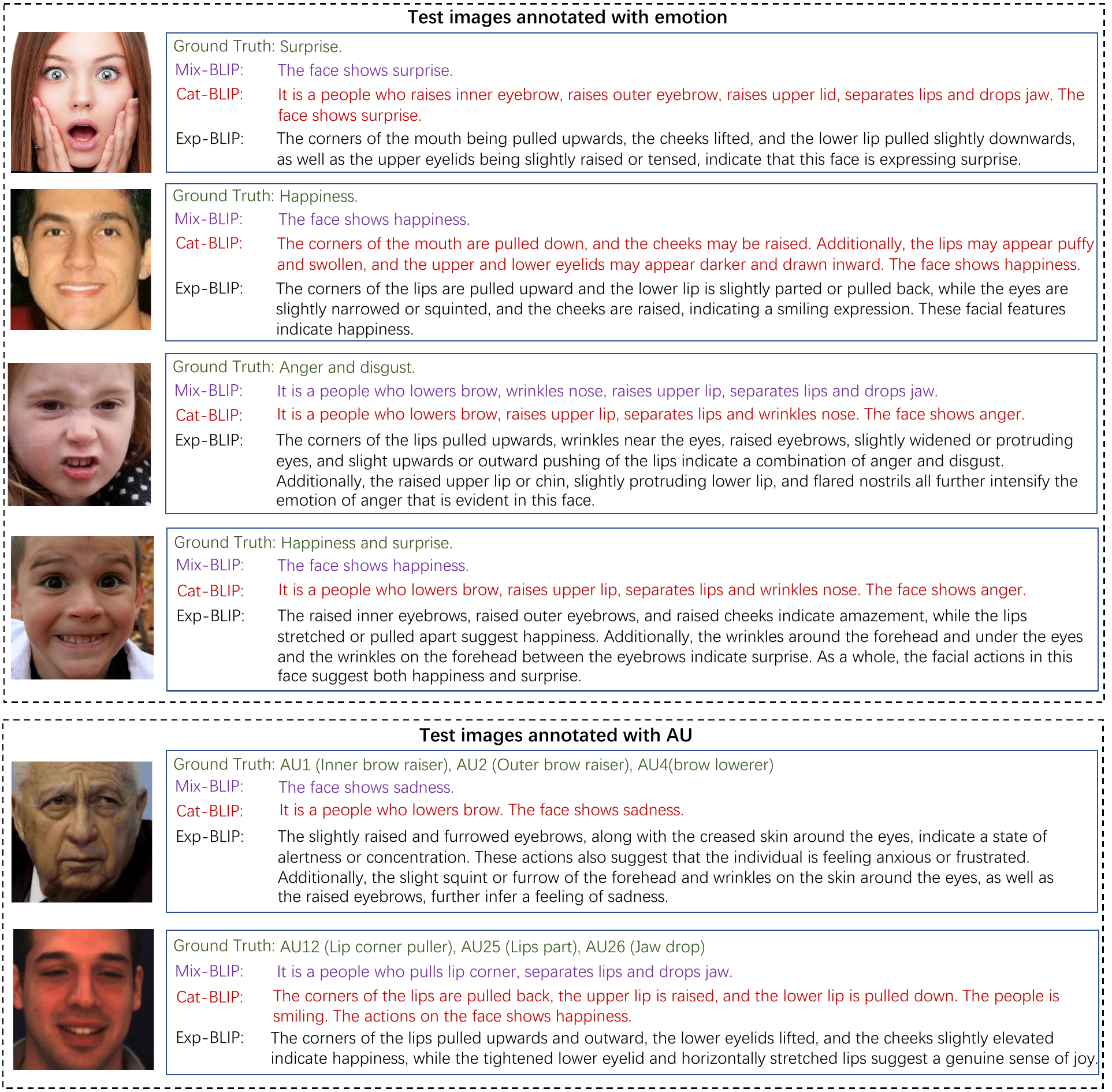}
    \vspace{-2.2 em}
    \caption{Examples of the generated captions by different models with ViT-G/OPT-6.7B}
    \label{fig: fused examples}
\end{figure}

\begin{figure}[!htbp]
    \centering
    \includegraphics[width=1.0\textwidth]{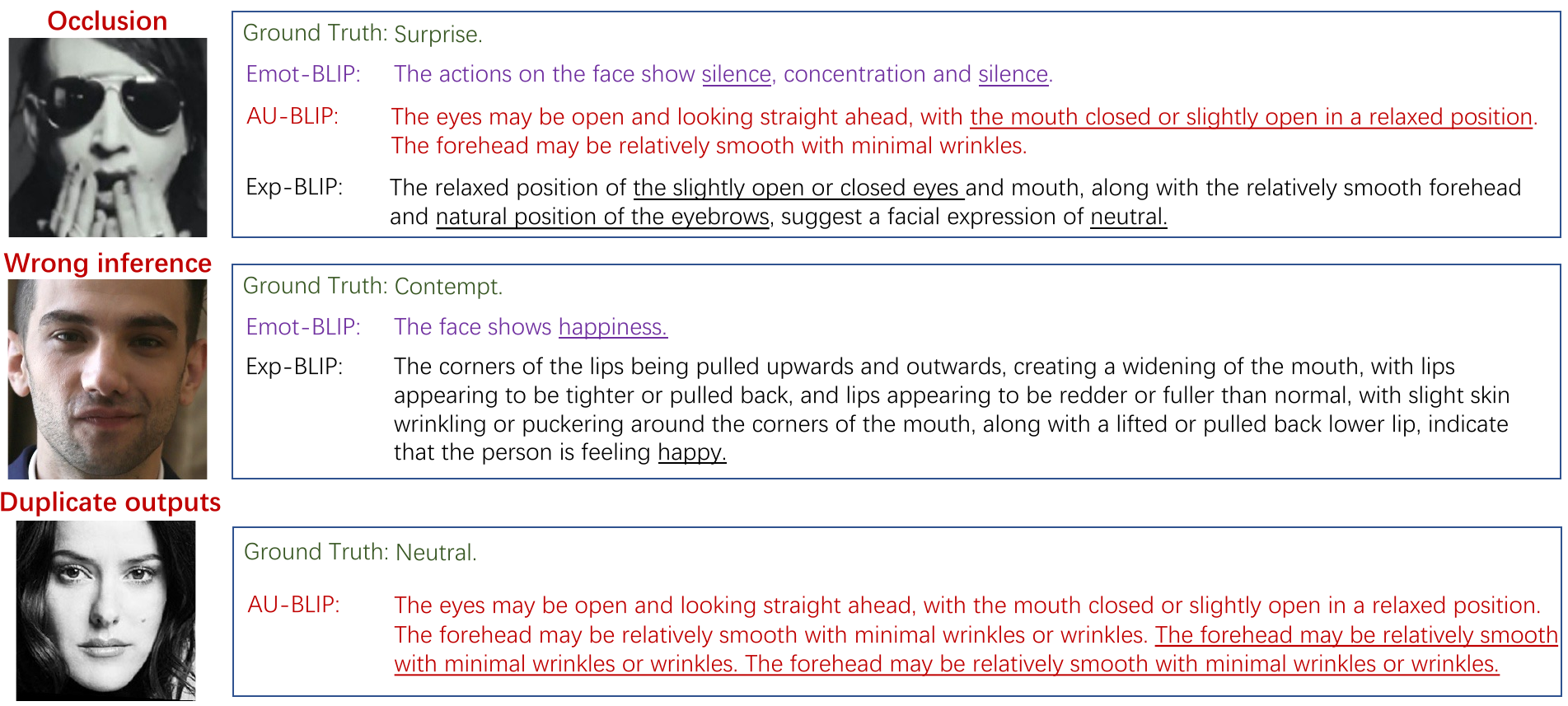}
    \vspace{-2.2 em}
    \caption{Bad cases of the generated captions for emotion-annotated images by different models with ViT-G/OPT-6.7B }
    \label{fig: bad examples}
\end{figure}

\section{Examples}
Due to the length constraints of the main text, we are unable to provide examples of Mix-BLIP, Cat-BLIP and Exp-BLIP in the main text. In order to intuitively present the difference of the outputs of Mix-BLIP, Cat-BLIP and Exp-BLIP, We show several examples of the three models' outputs in Figure~\ref{fig: fused examples}.
It can be observed that Mix-BLIP describes only one respect of AU and emotion randomly.
Cat-BLIP describes both AU and emotion while no connection between them.
Exp-BLIP describes not only both facial actions and emotions, but also the inferential relation between them.

Although our models demonstrate promising potential in describing facial expressions, they occasionally produce inaccurate results. Figure~\ref{fig: bad examples} illustrates three kinds of bad cases of our models and the bad sentences are underlined.
The first case is caused by large area of facial occlusion. This occlusion restricts the ability of image encoder to extract detail facial information, resulting a wrong description for facial expressions.
The second case is the wrong inference. Our Emot-BLIP and Exp-BLIP infer different emotions from the ground truth, which may be due to the limited capabilities of QFormer and the language model (OPT). Although labeled as contempt, this picture also slightly shows happiness as we predict.
The third case is generating duplicate outputs. The outputs of our models inherit the drawbacks of language model. OPT model sometimes tends to generate repeated and syntax error sentences, which will appear in the outputs of our models.

\vspace{-0.8 em}
\section{Hyperparameters}
\vspace{-0.5 em}
Table~\ref{tab:hp vit} and Table~\ref{tab:hp opt} show the Hyperparameters of image encoders and LLMs used in the main text, seperately. ViT-Base was trained on AffectNet dataset for 200 epochs under the training framework of MAE and a cosine
learning rate decay with a peak learning rate of 1e-4 was adopted. The architecture of Q-Former in Figure 3 in the main text is BERT-Base, with cross-attention layer inserted in each block. FFN means Feed Forward Network in the transformer block. The 
Hyperparameters of BERT-Base is listed in Table~\ref{tab:hp bert}. Table~\ref{tab:hp bert} shows the Hyperparameters of stage2 of fine-tuning BLIP-2 introduced in Sec. 3.2 of the main text.

\begin{table}[!htbp]
    \caption{Details of image encoder model (Vision Transformer) variants.}
    \label{tab:hp vit}
    \centering
    \scriptsize%
    \setlength{\tabcolsep}{4pt}%
    \renewcommand{\arraystretch}{1.2}%
    \vspace{0.2 em}
    \begin{tabular}{l|ccccc}
        \hline
        Model & Layers & Hidden size D & MLP size &  Heads & Params  \\
        \hline
        ViT-Base(ViT-B) & 12 & 768 & 3072 & 12 & 86M \\
        ViT-Giant(ViT-G) & 39 & 1408 & 6144 & 16 & 1843M \\
        \hline
    \end{tabular}
    \vspace{-2.2 em}
\end{table}

\begin{table}[!htbp]
    \caption{Hyperparameters for Large Language Model OPT2.7/OPT6.7 architecture.}
    \label{tab:hp opt}
    \centering
    \scriptsize%
    \setlength{\tabcolsep}{4pt}%
    \renewcommand{\arraystretch}{1.2}%
    \vspace{0.2 em}
    \begin{tabular}{l|cc}
        \hline
        LLM & $OPT_{2.7B}$ & $OPT_{6.7B}$   \\
        \hline
        FFN dim & 10240 & 16384  \\
        Hidden size & 2560 & 4096 \\
        Word embedding project dim & 2560 & 4096 \\
        Attention heads &\multicolumn{2}{c}{32} \\
        Layers & \multicolumn{2}{c}{32} \\
        Dropout & \multicolumn{2}{c}{0.1} \\
        Max position embeddings & \multicolumn{2}{c}{2048} \\
        \hline
    \end{tabular}
    \vspace{-2.2 em}
\end{table}

\begin{table}[!htbp]
    \caption{Hyperparameters for BERT-Base architecture.}
    \label{tab:hp bert}
    \centering
    \scriptsize%
    \setlength{\tabcolsep}{4pt}%
    \renewcommand{\arraystretch}{1.2}%
    \vspace{0.2 em}
    \begin{tabular}{l|c}
        \hline
        Model & $BERT_{Base}$    \\
        \hline
        Hidden size & 768 \\
        Intermediate size & 3072 \\
        Attention heads & 12 \\
        Layers & 12 \\
        max position embeddings & 512 \\
        \hline
    \end{tabular}
    \vspace{-2.2 em}
\end{table}

\begin{table}[!htbp]
    \caption{Hyperparameters for fine-tuning BLIP-2 with ViT-B/ViT-G on image captioning.}
    \label{tab:train config}
    \centering
    \scriptsize%
    \setlength{\tabcolsep}{4pt}%
    \renewcommand{\arraystretch}{1.2}%
    \vspace{0.2 em}
    \begin{tabular}{l|cc}
        \hline
        LLM & $OPT_{2.7B}$ & $OPT_{6.7B}$    \\
        \hline
        Fine-tuning epochs & \multicolumn{2}{c}{20} \\
        Warmup steps & \multicolumn{2}{c}{1000} \\
        Learning rate & \multicolumn{2}{c}{1e-5} \\
        Batch size & \multicolumn{2}{c}{32} \\
        AdamW $\beta$ & \multicolumn{2}{c}{(0.9,0.999)} \\
        Weight decay & \multicolumn{2}{c}{0.05} \\
        Drop path & \multicolumn{2}{c}{0} \\
        Image resolution & \multicolumn{2}{c}{224} \\
        Prompt & \multicolumn{2}{c}{""} \\
        Inference beam size & \multicolumn{2}{c}{5} \\
        Layer-wise learning rate decay for ViT & 1 & 0.95\\
        \hline
    \end{tabular}
    \vspace{-2.2 em}
\end{table}

\end{document}